\pdfoutput=1
\newif\ifbuildsupplement
\ifdefined\BuildSupplement\buildsupplementtrue\fi
\documentclass[10pt,twocolumn,letterpaper]{article}
\newif\ifarxivversion
\arxivversiontrue
\ifarxivversion
  \usepackage[pagenumbers]{cvpr}
  \makeatletter
  \providecommand{\@LN@col}[1]{}
  \patchcmd{\@maketitle}{\vskip .375in}{\vskip 0pt}{}{}
  \patchcmd{\@maketitle}{\vspace*{24pt}}{\vspace*{10pt}}{}{}
  \patchcmd{\@maketitle}{\vspace*{12pt}}{\vspace*{6pt}}{}{}
  \makeatother
\else
  \usepackage[review]{cvpr}
\fi

\usepackage{amsmath,amssymb,graphicx,booktabs,multirow,listings,capt-of}
\usepackage{algorithm,algpseudocode}
\definecolor{cvprblue}{rgb}{0.21,0.49,0.74}
\usepackage{xr-hyper}
\usepackage[pagebackref,breaklinks,colorlinks,allcolors=cvprblue]{hyperref}
\def\confName{CVPR}
\def\confYear{2027}
\newcommand{\miss}{\textemdash}
\ifbuildsupplement
\else
  \ifarxivversion\else
  \fi
\fi
\title{Cecilia: Towards Layout-Inclusive Optical Music Recognition\ifbuildsupplement\\[0.4em]\large Supplementary Material\fi}
\ifarxivversion
  \usepackage{marvosym}
  \ifbuildsupplement
    \author{}
    \hypersetup{pdfauthor={}}
  \else
  \author{Congren Dai\textsuperscript{1,2},
    Zhancheng Guo\textsuperscript{1,2},
    Xiaobing Li\textsuperscript{1},
    and Maosong Sun\textsuperscript{1,2\Letter}\\[5pt]
    \textsuperscript{1}Central Conservatory of Music\quad
    \textsuperscript{2}Tsinghua University\\
    {\tt\small congren.dai@mail.ccom.edu.cn\quad sms@tsinghua.edu.cn}}
  \hypersetup{pdfauthor={Congren Dai, Zhancheng Guo, Xiaobing Li, Maosong Sun}}
  \AddToHook{cmd/maketitle/after}{%
    \begingroup%
    \footnotetext[1]{Corresponding author.}\endgroup}
  \fi
\else
  \ifbuildsupplement\author{}\else\author{Anonymous submission}\fi
\fi
\ifbuildsupplement
  \makeatletter
  \patchcmd{\@maketitle}{Anonymous \confName~submission}{}{}{}
  \makeatother
\fi
\ExplSyntaxOn
\cs_new_protected:Npn \CeciliaSupplementRef #1
  {
    \tl_set:Nn \l_tmpa_tl {#1}
    \regex_match:nnTF {\AMain-} {#1}
      {
        \regex_replace_once:nnN {\AMain-} {} \l_tmpa_tl
        \exp_args:NV \CeciliaOriginalRef \l_tmpa_tl
      }
      { \CeciliaOriginalRef {SM-#1} }
  }
\ExplSyntaxOff

\begin{document}
\ifbuildsupplement
  \raggedbottom
  \maketitle
\setcounter{section}{0}
\renewcommand{\thesection}{\Alph{section}}
\renewcommand{\thetable}{S\arabic{table}}

\section{Token-length analysis}
We apply the same frozen Cecilia-OMR tokeniser to full-score ABC, literal-tree Cecilia and the corresponding MusicXML. Table~\ref{tab:tokens} reports population standard deviations ($\mathrm{ddof}=0$). The comparison contains 1,000 unique, successfully converted scores: 895 OpenScore Lieder and 105 String Quartets. All three formats must convert successfully for a score to enter the paired sample.

Token means for the complete score differ from the sum of split training-unit tokens because segmentation introduces headers, boundary state and special tokens.

\begin{table}[!htbp]
\centering\footnotesize
\begin{tabular}{@{}lrrr@{}}
\toprule
Format & Mean & Std. dev. & Median\\
\midrule
ABC & 11,175.86 & 20,413.60 & 5,626.5\\
Cecilia & 12,702.02 & 30,126.33 & 4,702.0\\
MusicXML & 352,228.32 & 681,560.15 & 156,222.0\\
\bottomrule
\end{tabular}
\caption{Token lengths for 1,000 paired complete scores under one fixed tokeniser. The statistics describe representation length under the common tokeniser.}
\label{tab:tokens}
\end{table}

The mean length of Cecilia is 1.14 times that of ABC while retaining the encoded score fields. The mean length of MusicXML is 27.73 times that of Cecilia. The token-length figure shows empirical histograms and descriptive Gaussian fits in log-ten token space. Its mean and standard-deviation annotations refer to raw token counts. The Gaussian curves summarise log-space shape, while raw lengths retain long right tails. A tokeniser trained separately for XML or ABC may change these ratios.

\section{Representation and target construction}
\subsection{Canonical-tree preservation}
Canonicalisation removes the \texttt{encoding}, \texttt{source}, \texttt{relation} and \texttt{miscellaneous} children of \texttt{identification}. Instrument identifiers are standardised consistently; device routing, MIDI channel, instrument-name playback overrides, volume, pan and elevation are removed. Musical notes, lyrics, title and credit text, instrument names, tempo, layout and \texttt{print-object="no"} elements remain. The structural pass normalises attribute order, line endings, formatting whitespace and negative-zero attributes. Algorithm~\ref{Main-alg:codec} decodes directly to the retained tree before any renderer-specific export transformations.

The decoder checks the literal-tree framing and complete consumption of the input, as specified in Algorithm~\ref{Main-alg:codec}. The generic form covers registered structures outside the specialised vocabulary. Horizontal differences use six-decimal rounding and reset at part and system boundaries. For example, positions 120, 145 and 170 are represented by 120, 25 and 25. A vertical baseline advances by five tenths per diatonic step; its coordinate is omitted only when the original value is reproduced exactly. An absence flag preserves a missing coordinate. Lyrics retain their explicit coordinates.

\subsection{Canonical-tree round-trip validation}
\label{sec:codec_roundtrip}
We test the literal-tree representation predicted by Cecilia-OMR on the same 1,000 complete scores used in the sequence-length comparison: 895 OpenScore Lieder and 105 String Quartets. Each canonical XML tree is encoded as a literal-tree sequence and reconstructed with its reader, without renderer-specific export transformations. The comparison includes element names, attribute keys and their string values, text, tails and ordered children. Canonicalisation sorts attributes, normalises line endings and negative-zero attributes, and removes whitespace-only formatting text.

All 1,000 reconstructed trees match their canonical inputs exactly, with zero exceptions and no numerical differences. This test evaluates the literal-tree encoder and reader directly, rather than only tokenisation or assembly of stored segments. The corpus contains 1,095,973 note horizontal-position attributes, all with two decimal places. Other numeric attributes have up to six decimal places.

The result is empirical for this corpus. Horizontal note differences are rounded to six decimal places. In a precision test, successive horizontal positions of 10.1234567 and 20.6543219 reconstruct as 10.1234567 and 20.6543219.

\subsection{Tokenisation and target admission}
Tokenisation uses frozen byte-level BPE with 8,192 vocabulary entries, no text normaliser and no regular-expression pre-tokeniser. The byte vocabulary encodes Unicode lyric text. A segment's 8,192-token budget includes beginning and end tokens. Training examples preserve complete authored measures. Boundary-state carriers contain the declarations needed for an independently decoded segment and are removed during assembly.

\section{Architecture and optimisation}
\subsection{Network and trainable parameters}
Table~\ref{tab:architecture} specifies the dimensions of the model in Section~\ref{Main-sec:method}. Decoder blocks use pre-normalisation and residual connections; text self-attention uses rotary positions. The visual encoder and projector retain the full patch grid.

\begin{table}[!htbp]
\centering\footnotesize
\begin{tabular}{@{}ll@{}}
\toprule
Component & Configuration\\
\midrule
Input & RGB, padded $1024\times1024$\\
Vision encoder & 12 blocks, width 768, patch 16\\
Memory per image & $4096\times2048$\\
Patch projector & $768\rightarrow2048\rightarrow2048$\\
Decoder & 28 blocks, width 2048\\
Attention & 16 heads, head width 128\\
Feed-forward width & 8192\\
Vocabulary & 8192; tied embedding/output\\
Segment conditions & 128 learned embeddings\\
Unit conditions & Metadata and music\\
Segment-count head & 128 classes\\
Maximum segment length & 8192 including BOS/EOS\\
\bottomrule
\end{tabular}
\caption{The active Cecilia-OMR architecture.}
\label{tab:architecture}
\end{table}

Supervised learning and policy updates optimise the visual encoder, projector, decoder, conditioning embeddings and segment-count head. GRPO uses a separately frozen copy as its reference policy.

Class $c$ of the segment-count head represents $c+1$ segments. The vocabulary projection uses FP32 computation, while the surrounding forward computation uses BF16. Cross-attention keys and values are cached for all segments of a unit.

\subsection{Training objectives and exposure}
\label{sec:optimisation}
The supervised objective in Section~\ref{Main-sec:method} uses the fixed token normaliser $C=1295.1014$ and count-classification weight 0.1. The following settings specify the subsequent policy refinement.

The GRPO stage starts from the mixed-data SFT model selected after 8,662 continuation updates. The policy and frozen reference initialise from identical SFT weights, and GRPO starts with fresh AdamW state. It samples four candidates for each work and uses two groups per optimiser batch across eight GPUs.

The sampled and replayed policies both use temperature 0.7. The optimiser uses learning rates $10^{-7}$ for non-visual parameters and $10^{-8}$ for the visual encoder, AdamW betas $(0.9,0.95)$, epsilon $10^{-8}$ and zero weight decay. The clipped policy ratio uses $\epsilon=0.2$ and the KL coefficient is 0.01.

Reward notation and format validity follow the definition in Section~\ref{Main-sec:method}. Format validity divides the number of valid metadata and music observation units by their fixed total. Unresolved scoring failures pause the affected optimisation group until scoring is available.

Within each four-sample group, advantages subtract the group mean and divide by the larger of the population reward standard deviation and 0.01. A standard deviation at most $10^{-8}$ skips the group entirely, including its KL term. The replay objective retains the fixed supervised token normaliser multiplied by the number of observed units, with fixed relative weights for metadata and music. 

The policy samples segment counts and tokens. Their replayed log probabilities define the group-relative update; reference notation supplies the reward. 

\subsection{Ground-truth-free decoding and recovery}
\label{sec:recovery}
Grammar-prefix decoding checks every proposed token against the running literal-tree reader state, including string encoding, numeric syntax and matching delimiters. The highest-scoring admissible continuation is selected. A greedy end-of-sequence decision is retained even when premature, and that candidate remains incomplete. Candidate selection uses native XML availability.

Recovery retains complete parsed subtrees and discards malformed trailing content. Metadata recovery combines compatible predicted roots and reconstructs part declarations from predicted identifiers. Music recovery uses divisions supplied by valid boundary carriers or compatible adjacent segments of the same part; duration-based inference is accepted only when the available constraints agree. These rules use generated content, and missing musical content remains absent. OMR-NED evaluates the resulting exported score.

The boundary carrier stores a positive divisions value for a contiguous run of measures from one part. Standalone export inserts this value into the first measure and removes the private carrier; full-score assembly removes the carrier while retaining the authored measures. It does not carry clefs, key signatures or time signatures.

\section{Evaluation}
\subsection{Reference scope}
The official OpenScore conditions retain their released reference XML~\cite{yang2026legato}. Canonicalisation for training does not modify these evaluation references. The additional Lieder references retain both the vocal staff and its lyrics.

CVC-MUSCIMA~\cite{fornes2012cvcmuscima} uses a work-disjoint split: two of its 20 musical scores are held out for testing, comprising 100 handwritten pages, while the remaining scores provide training data. Experts transcribe these scores into MusicXML, which we will release for further research.

\subsection{Baseline representations and inference}
The ABC and direct-XML baseline outputs described in the main paper are converted and rendered through the same downstream tools.

For the direct-XML condition, requests use the model identifier \texttt{gpt-6-astra} through \texttt{api.gpt.ge/v1/chat/completions}. Requests set temperature to zero and the output limit to 64,000 tokens. Each request contains one user message with the prompt and a PNG image.

Failed or incomplete requests are retried. Outputs are selected by the first completed response, without consulting reference scores. The following is the XML transcription prompt:

\begin{quote}\footnotesize
You are an optical music recognition system. The image is one page of printed sheet music.
Transcribe the full page into MusicXML.

Requirements:

Output one complete MusicXML score-partwise document, with a part-list and parts, including the printed title, meter, key and tempo, and valid divisions and note durations.

For a score with more than one staff, declare the corresponding parts, staves and voices in MusicXML, group them to match the printed bracing, and assign each note to its correct voice and staff.

Transcribe every staff, every system and every measure that is printed on this page, in score order. Do not invent music that is not printed, and do not skip printed music.

Include pitches, rests, durations, bar lines, clefs, key signatures, time signatures, accidentals, beams, ties, slurs, articulations, dynamics, tempo and other text directions.

If the page has sung text, write it in lyric elements attached to the notes of the voice that sings it. If the page has no sung text, do not add any lyric elements.

Output MusicXML only. No explanation, no commentary, no markdown code fences.
\end{quote}

For the ABC API baselines, the available reporting records do not establish immutable model snapshots or complete prompt and retry specifications. The model names in the results should therefore be read as reported service names. A complete accounting of billed cost and end-to-end latency is unavailable across the API runs. We report recognition accuracy, without making a cost or speed comparison between these services.

Legato 2 supplies published scores rather than prediction files, so its coverage is restricted to the four reported OMR-NED conditions.

\subsection{Music and text metrics}
\label{sec:metrics}
OMR-NED compares each target-page prediction with its corresponding page reference. Each domain sums the page-level notation edits and GT-plus-prediction notation sizes before division. The four-condition OpenScore macro is the equal-weight mean of domain OMR-NED values.

CER, SER and LER use the same deterministic XML-to-ABC conversion for prediction and reference. Their units are characters, lexical tokens including punctuation, and text lines. For each domain we divide the summed edit distance by the summed reference-unit count and multiply by 100. Scores can exceed 100 because insertions are not clipped. Their units are defined by this conversion.

TEDn compares notation trees derived from XML using rational durations. All predicted and reference parts enter the comparison, and the full authored measure sequences remain available to the metric. The edit cost for each page is divided by the deletion cost of its complete reference tree and multiplied by 100; the domain score averages these page percentages over its fixed sample count. An empty model output therefore contributes 100\%, while insertions can produce larger values. The tree representation excludes the lyric and layout fields measured separately. Scoring-resource failures remain unresolved until computation completes. Lyric-text CER is computed separately from lyric text extracted from XML and is distinct from ABC-based CER.

\subsection{Fixed-sample failure accounting}
All pages remain in their evaluation domain. Admissibility is metric-specific: for example, a prediction that supports notation comparison but not ABC conversion retains its OMR-NED and receives an empty prediction only for the affected text metrics. For empty predictions, notation edit counts equal the reference notation count and the prediction count is zero. For CER, SER and LER, all reference characters, lexical tokens or lines are deletions; the empty prediction contains zero lines. We aggregate these counts with successful pages before division. Successful error rates above 100\% remain unchanged.

Reference and evaluator failures remain separate from model errors. Table~\ref{tab:failures} gives the metric-specific counts.

\begin{table*}[!t]
\centering\footnotesize
\begin{tabular}{@{}lrrrrr@{}}
\toprule
Condition & $N$ & \multicolumn{2}{c}{Cecilia-OMR} & \multicolumn{2}{c}{GPT-6 Astra (XML)}\\
 & & Invalid & ABC-only & Invalid & ABC-only\\
\midrule
Rendered OpenScore String Quartets & 252 & 11 & 0 & 52 & 0\\
Camera OpenScore String Quartets & 252 & 22 & 0 & 56 & 0\\
Rendered OpenScore Lieder & 64 & 5 & 1 & 5 & 2\\
Camera OpenScore Lieder & 64 & 6 & 0 & 5 & 1\\
Rendered OpenScore Lieder with lyrics & 64 & 7 & 0 & 6 & 0\\
Camera OpenScore Lieder with lyrics & 64 & 13 & 0 & 5 & 0\\
CVC-MUSCIMA & 100 & 0 & 1 & 15 & 0\\
\bottomrule
\end{tabular}
\caption{Failure counts for notation and ABC-based text evaluation. Invalid denotes an unusable XML prediction under the notation evaluator. ABC-only denotes an additional model-structure failure in XML-to-ABC conversion whose notation score remains available. Counts record which predictions each metric can evaluate. All failure pages remain in the fixed evaluation sets.}
\label{tab:failures}
\end{table*}

The sequence metrics further distinguish content recovery from tree-level similarity. Their shared XML-to-ABC conversion exposes differences in characters, lexical units and lines under one deterministic representation. A low notation distance can therefore coexist with substantial line error when the predicted music is organised differently in the converted text. We report the conversion-based line error separately from the system-break F1 of final layout: the former concerns a symbolic text sequence, while the latter concerns where the rendered systems divide the score. Together, the two measures distinguish symbolic line structure from rendered system geometry.

\subsection{Final-layout metrics}
The reference and predicted MusicXML use the same MuseScore 3.2.3 renderer and fonts. Geometry is extracted from final SVG, positional exports and PDF page dimensions. Coordinates are normalised within the rendered page. The comparison measures agreement between the final layouts rendered from the two XML files.

System-break F1 compares system boundaries after alignment of logical measures. Page-break F1 compares page boundaries. Measures align by musical order; staffs align within matched systems; text words use normalised text and order. For an object category with $n_G$ reference and $n_P$ predicted objects, matched IoUs are summed and divided by $\max(n_G,n_P)$. Unmatched objects contribute zero. NPE uses centre displacement divided by the page diagonal, clipped to one; unmatched or cross-page objects contribute one, with the same object-count denominator. These rules include both matched geometry and omitted or extra objects.

Lyric-line initialisation is identical for all rendered predictions and references, including zero-length lyric segments. The notation, lyric content and geometric attributes are preserved before the common layout comparison.

The main geometric comparison evaluates Cecilia-OMR against Astra XML. It measures predicted layout instructions and automatic layout under one renderer. The other baselines enter the musical evaluation and qualitative comparison.

\section{Complete recognition and layout results}
Tables~\ref{tab:complete_secondary} and~\ref{tab:complete_geometry} give the numerical values and availability of results plotted in the main paper (Figure~\ref{Main-fig:metrics}). They preserve the complete domain sizes, common conversions and final-layout definitions described above. The geometric comparison is restricted to Cecilia-OMR and the direct-XML Astra condition, while all six systems with available predictions enter the secondary recognition table.

\begin{table*}[!t]
\centering\fontsize{7.5}{9}\selectfont\renewcommand{\arraystretch}{1.14}
\setlength{\tabcolsep}{1.5pt}
\begin{tabular*}{\textwidth}{@{\extracolsep{\fill}}ll*{7}{c}@{}}
\toprule
\multirow{3}{*}[-0.5ex]{Metric} & \multirow{3}{*}[-0.5ex]{Model} & \multicolumn{3}{c}{\textbf{Rendered}} & \multicolumn{3}{c}{\textbf{Camera}} & \textbf{Handwritten}\\
\cmidrule(lr){3-5}\cmidrule(lr){6-8}\cmidrule(l){9-9}
 & & String Quartets & Lieder & Lieder with lyrics & String Quartets & Lieder & Lieder with lyrics & CVC-MUSCIMA\\
 & & $N=252$ & $N=64$ & $N=64$ & $N=252$ & $N=64$ & $N=64$ & $N=100$\\
\midrule
CER (\%) $\downarrow$ & Cecilia-OMR & 41.71 & 38.08 & 41.16 & 41.72 & 57.71 & 51.91 & 40.53\\
 & GPT-6 Astra & 35.86 & 44.58 & 55.76 & 38.89 & 46.37 & 55.25 & 31.21\\
 & GPT-6 Astra (XML) & 54.53 & 61.30 & 50.90 & 59.23 & 65.18 & 53.49 & 41.59\\
 & Claude Fable 5.1 & 57.33 & 60.59 & 57.04 & 59.70 & 62.18 & 58.27 & 52.07\\
 & Gemini 4 Argon & 52.29 & 50.21 & 46.36 & 55.37 & 52.56 & 50.00 & 33.01\\
 & Legato 1 & 41.65 & 47.06 & 53.89 & 60.06 & 55.08 & 62.96 & 118.31\\
\midrule
SER (\%) $\downarrow$ & Cecilia-OMR & 41.18 & 39.57 & 42.87 & 42.37 & 55.29 & 54.24 & 48.66\\
 & GPT-6 Astra & 41.01 & 47.77 & 58.93 & 43.77 & 49.95 & 58.53 & 40.85\\
 & GPT-6 Astra (XML) & 54.66 & 62.86 & 53.60 & 59.57 & 66.59 & 56.22 & 43.25\\
 & Claude Fable 5.1 & 67.04 & 66.25 & 63.88 & 69.96 & 67.47 & 64.59 & 65.55\\
 & Gemini 4 Argon & 60.49 & 55.88 & 54.06 & 61.66 & 58.77 & 57.61 & 42.34\\
 & Legato 1 & 44.71 & 48.53 & 56.26 & 65.10 & 58.92 & 65.16 & 116.30\\
\midrule
LER (\%) $\downarrow$ & Cecilia-OMR & 70.02 & 63.72 & 67.39 & 94.31 & 84.97 & 77.51 & 60.40\\
 & GPT-6 Astra & 73.10 & 66.30 & 76.16 & 74.09 & 65.38 & 75.16 & 60.33\\
 & GPT-6 Astra (XML) & 78.38 & 81.55 & 75.88 & 82.21 & 85.67 & 82.46 & 72.74\\
 & Claude Fable 5.1 & 74.52 & 71.09 & 72.66 & 76.02 & 72.30 & 75.19 & 63.98\\
 & Gemini 4 Argon & 77.77 & 68.84 & 67.76 & 88.62 & 69.91 & 71.38 & 58.29\\
 & Legato 1 & 81.86 & 66.00 & 78.33 & 84.45 & 70.53 & 86.68 & 112.97\\
\midrule
TEDn (\%) $\downarrow$ & Cecilia-OMR & 32.15 & 24.32 & 27.57 & 21.96 & 28.08 & 39.32 & 28.51\\
 & GPT-6 Astra & 23.95 & 61.07 & 58.78 & 23.66 & 57.42 & 53.83 & 15.88\\
 & GPT-6 Astra (XML) & 41.34 & 39.45 & 23.46 & 43.02 & 37.36 & 23.63 & 24.96\\
 & Claude Fable 5.1 & 42.76 & 70.91 & 61.47 & 49.12 & 71.34 & 66.22 & 37.67\\
 & Gemini 4 Argon & 34.21 & 58.95 & 52.67 & 36.70 & 61.81 & 56.15 & 19.33\\
 & Legato 1 & 23.69 & 62.59 & 61.48 & 45.04 & 70.46 & 72.53 & 89.01\\
\bottomrule
\end{tabular*}
\caption{Complete numerical recognition results corresponding to the main paper\textquotesingle s metric figure. Every value retains the fixed domain size. CER, SER and LER aggregate edits and reference units across the domain; TEDn averages complete-page normalised tree distances. Values above 100\% remain unclipped. String Quartets, Lieder and Lieder with lyrics are OpenScore subsets. Cecilia-OMR uses the selected GRPO model.}
\label{tab:complete_secondary}
\end{table*}

\begin{table*}[!t]
\centering\fontsize{7.5}{9}\selectfont\renewcommand{\arraystretch}{1.14}
\setlength{\tabcolsep}{1.5pt}
\begin{tabular*}{\textwidth}{@{\extracolsep{\fill}}ll*{7}{c}@{}}
\toprule
\multirow{3}{*}[-0.5ex]{Metric} & \multirow{3}{*}[-0.5ex]{Model} & \multicolumn{3}{c}{\textbf{Rendered}} & \multicolumn{3}{c}{\textbf{Camera}} & \textbf{Handwritten}\\
\cmidrule(lr){3-5}\cmidrule(lr){6-8}\cmidrule(l){9-9}
 & & String Quartets & Lieder & Lieder with lyrics & String Quartets & Lieder & Lieder with lyrics & CVC-MUSCIMA\\
 & & $N=252$ & $N=64$ & $N=64$ & $N=252$ & $N=64$ & $N=64$ & $N=100$\\
\midrule
System-break F1 $\uparrow$ & Cecilia-OMR & 0.6829 & 0.7311 & 0.6989 & 0.6513 & 0.4324 & 0.4140 & 0.4231\\
 & GPT-6 Astra (XML) & 0.4404 & 0.4832 & 0.6240 & 0.4445 & 0.4681 & 0.6159 & 0.4258\\
\midrule
Page-break F1 $\uparrow$ & Cecilia-OMR & 0.6534 & 0.7656 & 0.5729 & 0.1039 & 0.7344 & 0.3125 & 0.9700\\
 & GPT-6 Astra (XML) & 0.0913 & 0.6406 & 0.4062 & 0.0483 & 0.7031 & 0.3906 & 0.8400\\
\midrule
Measure IoU $\uparrow$ & Cecilia-OMR & 0.2340 & 0.3829 & 0.4017 & 0.0729 & 0.1473 & 0.1737 & 0.0556\\
 & GPT-6 Astra (XML) & 0.0193 & 0.1712 & 0.2809 & 0.0212 & 0.1898 & 0.2623 & 0.0912\\
\midrule
Staff IoU $\uparrow$ & Cecilia-OMR & 0.0240 & 0.1756 & 0.1172 & 0.0060 & 0.0668 & 0.0513 & 0.0936\\
 & GPT-6 Astra (XML) & 0.0001 & 0.0751 & 0.0682 & 0.0001 & 0.0657 & 0.0648 & 0.1236\\
\midrule
Text-word IoU $\uparrow$ & Cecilia-OMR & 0.0466 & 0.1043 & 0.0393 & 0.0066 & 0.0090 & 0.0058 & 0.0097\\
 & GPT-6 Astra (XML) & 0.0041 & 0.0307 & 0.0205 & 0.0010 & 0.0182 & 0.0131 & 0.0098\\
\midrule
Measure NPE $\downarrow$ & Cecilia-OMR & 0.3877 & 0.2733 & 0.3530 & 0.6008 & 0.3633 & 0.5010 & 0.5331\\
 & GPT-6 Astra (XML) & 0.7686 & 0.5427 & 0.4465 & 0.7581 & 0.5004 & 0.4610 & 0.3913\\
\midrule
Staff NPE $\downarrow$ & Cecilia-OMR & 0.3886 & 0.2557 & 0.3558 & 0.6608 & 0.3878 & 0.4923 & 0.3511\\
 & GPT-6 Astra (XML) & 0.8125 & 0.5436 & 0.4180 & 0.8042 & 0.4251 & 0.4287 & 0.3351\\
\midrule
Text-word NPE $\downarrow$ & Cecilia-OMR & 0.7896 & 0.6799 & 0.6707 & 0.9208 & 0.9150 & 0.8764 & 0.8662\\
 & GPT-6 Astra (XML) & 0.9326 & 0.7376 & 0.7081 & 0.9508 & 0.7381 & 0.7441 & 0.6747\\
\bottomrule
\end{tabular*}
\caption{Final-layout geometry results. The same reference and prediction XML pairs enter the system-break, page-break, overlap and normalised position measurements. Empty outputs retain their prescribed penalties. These values compare final rendered layouts, not registration to the photographed or handwritten input. String Quartets, Lieder and Lieder with lyrics are OpenScore subsets. Cecilia-OMR uses the selected GRPO model.}
\label{tab:complete_geometry}
\end{table*}

\section{Decoding strategy comparison}
The two strategies in Table~\ref{Main-tab:decoding_strategy} use the same Cecilia-OMR weights, ordered image windows, reference XML and output processing. Model parameters and the output projection use FP32, forward computation uses BF16 autocast, and TF32 is disabled. The comparison retains all 632 OpenScore benchmark pages, 128 Lieder with lyrics pages and 100 CVC-MUSCIMA pages. XML-availability selection uses the compiled candidate outputs to choose between unconstrained and grammar-prefix generation; reference XML enters only the subsequent evaluation. Both selected outputs receive the same deterministic recovery before evaluation.

\section{Predicted layout and automatic layout}
We compare two renderings of each Cecilia-OMR prediction on 48 fixed pages, with eight pages per condition. The first retains predicted layout instructions. The second removes page, system, staff and measure layout elements, explicit break and spacing attributes, measure widths, and position and curve-control coordinates. Both retain the same musical events, lyric attachments, part labels, scaling, fonts and appearance settings. An independent tree comparison verifies equality of all retained content. Both versions use the same MuseScore renderer and reference artifact. Table~\ref{Main-tab:layout_ablation} reports means over all 48 entries.

\section{Training-stage domain results}
\begin{table*}[!t]
\centering\small
\begin{tabular}{@{}lrrrrrr@{}}
\toprule
& \multicolumn{4}{c}{OpenScore benchmarks} & \multicolumn{2}{c}{Lieder with lyrics}\\
\cmidrule(lr){2-5}\cmidrule(l){6-7}
Stage & Rendered & Camera & Rendered & Camera & Rendered & Camera\\
 & Quartets & Quartets & Lieder & Lieder & & \\
\midrule
Base & .3615 & .5569 & .1308 & .4201 & .2478 & .5634\\
SFT & .2372 & .2359 & .2187 & .2336 & .2383 & .2732\\
GRPO & .2513 & .2307 & .1839 & .2270 & .2048 & .2862\\
\bottomrule
\end{tabular}
\caption{Domain-level OMR-NED for the training-stage comparison in Figure~\ref{Main-fig:scaling_demo}. Each Quartets condition contains 252 pages; each Lieder condition contains 64.}
\label{tab:model_variants}
\end{table*}

\section{Additional controlled evaluations}
\label{sec:additional_evaluations}
\subsection{Reward variants}
\label{sec:lyric_reward_ablation}
The reward variants initialise the policy and frozen reference from the same SFT model. Each uses 16 training groups, four candidates per group and seven optimiser updates. The original reward is $R_{\mathrm{original}}=0.95M+0.05F$, where $M$ is native notation accuracy and $F$ is format validity. The two additional variants use
\begin{align}
R_{\mathrm{lyric}}&=0.80M+0.15L+0.05F,\\
R_{\mathrm{format}}&=0.90M+0.10F.
\end{align}
The lyric variant applies to Lieder with lyrics; Quartets retain the original reward. Here, $L=\max(0,1-E_{\ell}/G_{\ell})$, with lyric character edits $E_{\ell}$ and reference character count $G_{\ell}$. Optimisation retains the original learning rate, temperature, training examples and group order. Each variant samples candidates from its current policy. Evaluation uses the same 48 target pages, with eight pages per condition. This small evaluation and short refinement support an exploratory comparison rather than a general conclusion about the best reward design.

Table~\ref{Main-tab:reward_main} reports the six notation scores. Table~\ref{tab:reward_lyrics} adds lyric-text CER for the two new variants. Each entry pools character edits and reference characters over eight pages.

\begin{table}[!htbp]
\centering\small
\begin{tabular}{@{}lrr@{}}
\toprule
Reward & Rendered (\%) & Camera (\%)\\
\midrule
Lyric & 12.2795 & 95.1220\\
Format & 11.7368 & 99.1516\\
\bottomrule
\end{tabular}
\caption{Lyric-text CER for the new GRPO reward variants. Each score pools eight pages; lower is better.}
\label{tab:reward_lyrics}
\end{table}

\subsection{Sampling temperature and candidate count}
\label{sec:temperature_candidates}
We compare temperatures $T\in\{0.3,0.7,1.0\}$ on 48 target pages, with eight pages from each of the six OpenScore conditions. Cecilia-OMR generates four independently seeded candidates at each temperature. A candidate pairs metadata and music from the same seed. The best-of-$N$ comparisons are nested: $N=1$ retains the first candidate, $N=2$ selects from the first two, and $N=4$ selects from all four.

Selection first prefers a complete pair with valid native XML, then maximises the mean temperature-adjusted log probability of its segment-count and token decisions. The denominator counts two unit-count decisions and all generated token decisions; BOS is excluded and EOS included. Ties retain the lower candidate index. If every candidate is invalid, the same probability rule selects an output whose invalid status is retained. Selection is performed separately at each temperature. The selected output receives the usual deterministic recovery. Reference notation is used only for evaluation. The generation budget permits up to 16 segments per unit and 8,192 tokens per segment; a sampled count above the segment cap produces an invalid candidate.

\begin{table*}[!t]
\centering\small
\setlength{\tabcolsep}{4pt}
\begin{tabular}{@{}lrrrrrrrrr@{}}
\toprule
& & \multicolumn{6}{c}{OMR-NED $\downarrow$} & \multicolumn{2}{c}{Lyric CER (\%) $\downarrow$}\\
\cmidrule(lr){3-8}\cmidrule(l){9-10}
$T$ & $N$ & RQ & CQ & RL & CL & RL+L & CL+L & Rendered & Camera\\
\midrule
$0.3$ & 1 & 0.3273 & 0.3881 & 0.1448 & 0.4248 & 0.2812 & 0.4434 & 46.68 & 72.96\\
$0.3$ & 2 & 0.3627 & 0.2999 & 0.1244 & 0.4662 & 0.2353 & 0.2400 & 39.82 & 68.61\\
$0.3$ & 4 & 0.3486 & 0.2342 & 0.1184 & 0.4277 & 0.2920 & 0.2863 & 44.64 & 71.05\\
$0.7$ & 1 & 0.3409 & 0.3580 & 0.2522 & 0.4111 & 0.3910 & 0.4259 & 48.91 & 65.64\\
$0.7$ & 2 & 0.3166 & 0.2763 & 0.1299 & 0.5651 & 0.4765 & 0.2777 & 52.65 & 75.50\\
$0.7$ & 4 & 0.3571 & 0.1438 & 0.0919 & 0.4854 & 0.3272 & 0.1968 & 19.00 & 74.23\\
$1.0$ & 1 & 0.4542 & 0.7486 & 0.4317 & 0.5280 & 0.6343 & 0.8120 & 59.91 & 96.71\\
$1.0$ & 2 & 0.3640 & 0.6352 & 0.1753 & 0.5308 & 0.5841 & 0.8250 & 57.19 & 99.15\\
$1.0$ & 4 & 0.3044 & 0.5394 & 0.1880 & 0.6805 & 0.4971 & 0.6342 & 56.78 & 90.99\\
\bottomrule
\end{tabular}
\caption{Sampling temperature and nested candidate-count results. Every domain contains the same eight target pages at all settings. RQ/CQ denote rendered/camera Quartets, RL/CL denote rendered/camera Lieder, and +L retains the vocal staff and lyrics. OMR-NED pools notation counts within each domain; lyric CER pools character counts. Candidate selection uses native XML validity and model probabilities, with references used only for evaluation.}
\label{tab:sampling_complete}
\end{table*}

The per-domain values in Table~\ref{tab:sampling_complete} supplement the macro comparison in Table~\ref{Main-tab:sampling_main}. All settings retain the same page and reference identities, with independent seeds reused across the nested candidate sets.

\section{The Figure 1 passage in three formats}
\label{sec:format_example}
The following listings give the complete passage displayed in Figure~\ref{Main-fig:representation}(a): the vocal part, measures~5--16, of Roger Quilter's \emph{The Night Piece}, together with its title and credit information. ABC, Cecilia and MusicXML describe the same twelve measures; these include the shorter examples in Figure~\ref{Main-fig:representation}(b). Cecilia and MusicXML retain the layout fields, while ABC expresses the musical events and lyrics. The final words ``of'' continue beyond the displayed passage in the original score. Long lines wrap for display.
\lstset{basicstyle=\ttfamily\fontsize{8}{10}\selectfont,breaklines=true,breakatwhitespace=false,literate={’}{{'}}1}

\subsection{ABC}
\begin{lstlisting}
X:1
T:To Julia, Op.8
T:The Night Piece
C:Roger Quilter
Z:Robert Herrick
Z:OpenScore (CC0)
L:1/8
M:2/4
I:linebreak $
K:C
V:1 treble nm="Clarinet in C"
V:1
 z2 z!p! E | A3/2 B/ A G | E2 E E | A3/2 B/ A G |$ %4
w: Her|eyes the glow- worm|lend thee, The|shoot- ing stars at-|
 E2 E E/ E/ | D D G G | D/ E/ D E G/ G/ |!<(! A c B G!<)! |$ %8
w: tend thee; And the|elves al- so, Whose|lit- tle eyes glow Like the|sparks of fire, the|
 (e>d c) G | (A>B A>B | G>A F>G | (E2 D)) A | %12
w: sparks, _ _ the|sparks _ _ _|_ _ _ _|* * of|

\end{lstlisting}

\subsection{Cecilia}
\begin{lstlisting}[firstline=2]
(|0K@e:4.0(|11(|12=#"To Julia, Op.8"))(|29=5)(|1O=#"The Night Piece")(|0L(|17=#"Transcribed from the IMSLP file(s) at https://imslp.org/wiki/Special:ReverseLookup/321919"@D:#"arranger")(|17=#"Roger Quilter"@D:#Z)(|17=#"Robert Herrick"@D:#"lyricist")(|1m=#"OpenScore (CC0)"))(|0M(|0N(|0O=6.7056)(|0P=40))(|0Q(|0R=1771.21)(|0S=1253.03)(|0D@D:#S(r=59.6517)(s=59.6517)(|0E=59.6517)(|0F=90.9091))(|0D@D:#T(r=59.6517)(s=59.6517)(|0E=59.6517)(|0F=90.9091)))(#"appearance"(#"line-widt
h"=1.8@D:#"light barline")(#"line-width"=5.5@D:#"heavy barline")(#"line-width"=5@D:#"beam")(#"line-width"=4.4@D:#1O)(#"line-width"=1.5@D:#1P)(#"line-width"=1@D:#"enclosure")(#"line-width"=1.1@D:#"ending")(#"line-width"=1@D:#"extend")(#"line-width"=1.6@D:#"leger")(#"line-width"=1.1@D:#2p)(#"line-width"=1.2@D:#"octave shift")(#"line-width"=2.1@D:#"slur middle")(#"line-width"=0.7@D:#"slur tip")(#"line-width"=1.1@D:#"staff")(#"line-width"=1.1@D:#"stem")(#"line-width"=2.1@D:#"tie middle")(#"line-widt
h"=0.7@D:#"tie tip")(#"line-width"=1@D:#"tuplet bracket")(#"line-width"=1.2@D:#"wedge")(#"note-size"=70@D:#U)(#"note-size"=70@D:#"grace")(#"note-size"=49@D:#"grace-cue"))(#"music-font"@N:#"Leland")(|0T@N:#"Edwin",O:10)(|0U@N:#"Edwin",O:10))(|0v@X:1(#"credit-type"=#"title")(|0t=#"To Julia, Op.8"@C:626.515152,B:1711.560455,O:14,Y:#P,Z:#R))(|0v@X:1(#"credit-type"=#"subtitle")(|0t=#"5. The Night Piece"@C:626.515152,B:1651.90882,O:20,Y:#P,Z:#R))(|0v@X:1(#"credit-type"=#"lyricist")(|0t=#"Robert Herric
k"@C:59.651667,B:1554.994599,Y:#r,Z:#V))(|0v@X:1(#"credit-type"=#Z)(|0t=#"Roger Quilter"@C:1193.378636,B:1554.994599,Y:#p,Z:#V))(|0v@X:1(#"credit-type"=#"rights")(|0t=#"OpenScore (CC0)"@C:626.515152,B:90.909091,O:9,Y:#P,Z:#V))(|0V(|0i@G:#"P1"(|0j=#"Clarinet in C"@R:#m)(6@G:#"P1-I1"(7=#"Clarinet")(#"instrument-sound"=#"wind.reed.clarinet"))(|0h@G:#"P1-I1")(8@G:#"P1-I1"(9=72))))(|0k@G:#"P1"[@A:5,E:332.3(w(|0l=4)(|02(|03=0))(1(2=2)(3=4))(|05(|06=#8)(|07=2)))(Y@K:#i(t(u(r=0)(s=0))(|00=125.05))){@C:6
1.36,B:-20!$4;&1;'3;}{@C:124.22,B:-20!$2;&1;'0;}<@I:#n(j(p@C:3.09,H:15(|0s)))(o@F:54.44)>{@C:62.11,B:-40:E4;$2;&1;'0;.1;/@C:6.5,B:-48.69,A:1,H:-30\o;_"Her";/}][@A:6,E:270.59{@C:-237.69:A4;$3;&1;'0;(U@C:34.55,B:-25).1;/@C:6.5,B:-48.69,A:1,H:-30\o;_"eyes";/}{@C:93.17:B4;$1;&1;'7;.2;/@C:6.5,B:-48.69,A:1,H:-30\o;_"the";/}{@C:41.40:A4;$2;&1;'0;.1;/@C:6.5,B:-48.69,A:1,H:-30\6;_"glow";/}{@C:62.11:G4;$2;&1;'0;.1;/@C:6.5,B:-48.69,A:1,H:-30\5;_"worm";/}][@A:7,E:260.24{@C:-196.68:E4;$4;&1;'3;.1;/@C:6.5,B:-
48.69,A:1,H:-30\o;_"lend";/}{@C:124.22:E4;$2;&1;'0;.1;/@C:8.81,B:-48.69,A:1,H:-30\o;_"thee,";/}{@C:62.11:E4;$2;&1;'0;.1;/@C:6.5,B:-48.69,A:1,H:-30\o;_"The";/}][@A:8,E:270.59{@C:-186.33:A4;$3;&1;'0;(U@C:34.55,B:-25).1;/@C:6.5,B:-48.69,A:1,H:-30\6;_"shoot";/}{@C:93.17:B4;$1;&1;'7;.1;/@C:6.5,B:-48.69,A:1,H:-30\5;_"ing";/}{@C:41.40:A4;$2;&1;'0;.1;/@C:6.5,B:-48.69,A:1,H:-30\o;_"stars";/}{@C:62.11:G4;$2;&1;'0;.1;/@C:6.5,B:-48.69,A:1,H:-30\6;_"at";/}][@A:9,E:323.99(Y@K:#i(t(u(r=0)(s=0))(|00=125.05))){@
C:61.36,B:-40:E4;$4;&1;'3;.1;/@C:6.5,B:-47.63,A:1,H:-30\5;_"tend";/}{@C:120.38:E4;$2;&1;'0;.1;/@C:9.02,B:-47.63,A:1,H:-30\o;_"thee;";/}{@C:60.19:E4;$1;&1;'7;.1;/@C:6.5,B:-47.63,A:1,H:-30\o;_"And";/}{@C:40.13:E4;$1;&1;'7;.1;/@C:6.5,B:-47.63,A:1,H:-30\o;_"the";/}][@A:10,E:264.46{@C:-260.17:D4;$2;&1;'0;.1;/@C:6.5,B:-47.63,A:1,H:-30\o;_"elves";/}{@C:60.20:D4;$2;&1;'0;.1;/@C:6.5,B:-47.63,A:1,H:-30\6;_"al";/}{@C:60.19:G4;$2;&1;'0;.1;/@C:9.02,B:-47.63,A:1,H:-30\5;_"so,";/}{@C:60.19:G4;$2;&1;'0;.1;/@C:6
.5,B:-47.63,A:1,H:-30\o;_"Whose";/}][@A:11,E:292.7{@C:-192.47:D4;$1;&1;'7;.1;/@C:6.5,B:-47.63,A:1,H:-30\6;_"lit";/}{@C:40.13:E4;$1;&1;'7;.1;/@C:6.5,B:-47.63,A:1,H:-30\5;_"tle";/}{@C:40.13:D4;$2;&1;'0;.1;/@C:6.5,B:-47.63,A:1,H:-30\o;_"eyes";/}{@C:60.19:E4;$2;&1;'0;.1;/@C:6.5,B:-47.63,A:1,H:-30\o;_"glow";/}{@C:60.19:G4;$1;&1;'7;.1;/@C:6.5,B:-47.63,A:1,H:-30\o;_"Like";/}{@C:40.13:G4;$1;&1;'7;.1;/@C:6.5,B:-47.63,A:1,H:-30\o;_"the";/}][@A:12,E:252.57<@I:#n(j(0@B:25,A:1,#"spread":13,D:#z))>{@C:-240.77
:A4;$2;&1;'0;.1;/@C:6.5,B:-47.63,A:1,H:-30\o;_"sparks";/}{@C:60.19:C5;$2;&1;'0;.2;/@C:6.5,B:-47.63,A:1,H:-30\o;_"of";/}{@C:60.20:B4;$2;&1;'0;.2;/@C:8.81,B:-47.63,A:1,H:-30\o;_"fire,";/}{@C:60.19:G4;$2;&1;'0;.1;/@C:6.5,B:-47.63,A:1,H:-30\o;_"the";/}<@I:#n(j(0@A:1,D:#g))>][@A:13,E:313.11(Y@K:#i(t(u(r=0)(s=0))(|00=125.05))){@C:61.36,B:-5:E5;$3;&1;'0;(U@C:77.85,B:-5).2;(I=#6@A:1)(N(l@#"bezier-x":28.28489,#"bezier-y":16.528239,A:1,D:#h))/@C:0.84,B:-48.4,A:1,H:-30\o;_"sparks,";(|0A)/}{@C:89.98:D5;$1;&
1;'7;.2;(I=#5@A:1)(I=#v@A:2)}{@C:39.99:C5;$2;&1;'0;.2;(N(l@#"bezier-x":-25.42371,#"bezier-y":20.660413,A:1,D:#g))}{@C:59.99:G4;$2;&1;'0;.1;/@C:6.5,B:-48.4,A:1,H:-30\o;_"the";/}][@A:14,E:273.79{@C:-239.28:A4;$3;&1;'0;(U@C:28.53,B:-25).1;(I=#6@A:1)(N(l@#"bezier-x":107.442625,#"bezier-y":38.184258,A:1,#"orientation":#"over",I:#n,D:#h))/@B:-48.4,A:1,H:-30\o;_"sparks";(|0A)/}{@C:89.98:B4;$1;&1;'7;.1;(I=#5@A:1)(I=#v@A:2)}{@C:39.99:A4;$3;&1;'0;(U@C:158.51,B:-25).1;(I=#6@A:1)}{@C:89.99:B4;$1;&1;'7;.1;(I
=#5@A:1)(I=#v@A:2)}][@A:15,E:275.07{@C:-218.68:G4;$3;&1;'0;(U@C:29.81,B:-25).1;(I=#6@A:1)}{@C:89.98:A4;$1;&1;'7;.1;(I=#5@A:1)(I=#v@A:2)}{@C:40.00:F4;$3;&1;'0;(U@C:159.79,B:-35).1;(I=#6@A:1)}{@C:89.98:G4;$1;&1;'7;.1;(I=#5@A:1)(I=#v@A:2)}][@A:16,E:271.75{@C:-223.28:E4;$4;&1;'3;.1;(N(l@#"bezier-x":28.229942,#"bezier-y":-20.35162,A:2,D:#h))}{@C:139.97:D4;$2;&1;'0;.1;(N(l@#"bezier-x":-103.85417,#"bezier-y":47.077243,A:1,D:#g)(l@#"bezier-x":-29.610089,#"bezier-y":-18.285533,A:2,D:#g))}{@C:59.99:A4;$2;
&1;'0;.1;/@C:6.5,B:-48.4,A:1,H:-30\o;_"of";/}]))

\end{lstlisting}

\subsection{MusicXML}
\begin{lstlisting}
<?xml version='1.0' encoding='utf-8'?>
<score-partwise version="4.0"><work><work-title>To Julia, Op.8</work-title></work><movement-number>5</movement-number><movement-title>The Night Piece</movement-title><identification><creator type="arranger">Transcribed from the IMSLP file(s) at https://imslp.org/wiki/Special:ReverseLookup/321919</creator><creator type="composer">Roger Quilter</creator><creator type="lyricist">Robert Herrick</creator><rights>OpenScore (CC0)</rights></identification><defaults
><scaling><millimeters>6.7056</millimeters><tenths>40</tenths></scaling><page-layout><page-height>1771.21</page-height><page-width>1253.03</page-width><page-margins type="even"><left-margin>59.6517</left-margin><right-margin>59.6517</right-margin><top-margin>59.6517</top-margin><bottom-margin>90.9091</bottom-margin></page-margins><page-margins type="odd"><left-margin>59.6517</left-margin><right-margin>59.6517</right-margin><top-margin>59.6517</top-margin><bottom-margin>90.9091</bottom-margin></p
age-margins></page-layout><appearance><line-width type="light barline">1.8</line-width><line-width type="heavy barline">5.5</line-width><line-width type="beam">5</line-width><line-width type="bracket">4.4</line-width><line-width type="dashes">1.5</line-width><line-width type="enclosure">1</line-width><line-width type="ending">1.1</line-width><line-width type="extend">1</line-width><line-width type="leger">1.6</line-width><line-width type="pedal">1.1</line-width><line-width type="octave shift">1.
2</line-width><line-width type="slur middle">2.1</line-width><line-width type="slur tip">0.7</line-width><line-width type="staff">1.1</line-width><line-width type="stem">1.1</line-width><line-width type="tie middle">2.1</line-width><line-width type="tie tip">0.7</line-width><line-width type="tuplet bracket">1</line-width><line-width type="wedge">1.2</line-width><note-size type="cue">70</note-size><note-size type="grace">70</note-size><note-size type="grace-cue">49</note-size></appearance><music-
font font-family="Leland" /><word-font font-family="Edwin" font-size="10" /><lyric-font font-family="Edwin" font-size="10" /></defaults><credit page="1"><credit-type>title</credit-type><credit-words default-x="626.515152" default-y="1711.560455" justify="center" valign="top" font-size="14">To Julia, Op.8</credit-words></credit><credit page="1"><credit-type>subtitle</credit-type><credit-words default-x="626.515152" default-y="1651.90882" justify="center" valign="top" font-size="20">5. The Night P
iece</credit-words></credit><credit page="1"><credit-type>lyricist</credit-type><credit-words default-x="59.651667" default-y="1554.994599" justify="left" valign="bottom">Robert Herrick</credit-words></credit><credit page="1"><credit-type>composer</credit-type><credit-words default-x="1193.378636" default-y="1554.994599" justify="right" valign="bottom">Roger Quilter</credit-words></credit><credit page="1"><credit-type>rights</credit-type><credit-words default-x="626.515152" default-y="90.909091"
 justify="center" valign="bottom" font-size="9">OpenScore (CC0)</credit-words></credit><part-list><score-part id="P1"><part-name print-object="no">Clarinet in C</part-name><score-instrument id="P1-I1"><instrument-name>Clarinet</instrument-name><instrument-sound>wind.reed.clarinet</instrument-sound></score-instrument><midi-device id="P1-I1" /><midi-instrument id="P1-I1"><midi-program>72</midi-program></midi-instrument></score-part></part-list><part id="P1"><measure number="5" width="332.3"><attri
butes><divisions>4</divisions><key><fifths>0</fifths></key><time><beats>2</beats><beat-type>4</beat-type></time><clef><sign>G</sign><line>2</line></clef></attributes><print new-system="yes"><system-layout><system-margins><left-margin>0</left-margin><right-margin>0</right-margin></system-margins><system-distance>125.05</system-distance></system-layout></print><note default-x="61.36" default-y="-20"><rest /><duration>4</duration><voice>1</voice><type>quarter</type></note><note default-x="185.58" d
efault-y="-20"><rest /><duration>2</duration><voice>1</voice><type>eighth</type></note><direction placement="above"><direction-type><dynamics default-x="3.09" relative-y="15"><p /></dynamics></direction-type><sound dynamics="54.44" /></direction><note default-x="247.69" default-y="-40"><pitch><step>E</step><octave>4</octave></pitch><duration>2</duration><voice>1</voice><type>eighth</type><stem>up</stem><lyric number="1" default-x="6.5" default-y="-48.69" relative-y="-30"><syllabic>single</syllab
ic><text>Her</text></lyric></note></measure><measure number="6" width="270.59"><note default-x="10.00" default-y="-25"><pitch><step>A</step><octave>4</octave></pitch><duration>3</duration><voice>1</voice><type>eighth</type><dot default-x="34.55" default-y="-25" /><stem>up</stem><lyric number="1" default-x="6.5" default-y="-48.69" relative-y="-30"><syllabic>single</syllabic><text>eyes</text></lyric></note><note default-x="103.17" default-y="-20"><pitch><step>B</step><octave>4</octave></pitch><dur
ation>1</duration><voice>1</voice><type>16th</type><stem>down</stem><lyric number="1" default-x="6.5" default-y="-48.69" relative-y="-30"><syllabic>single</syllabic><text>the</text></lyric></note><note default-x="144.57" default-y="-25"><pitch><step>A</step><octave>4</octave></pitch><duration>2</duration><voice>1</voice><type>eighth</type><stem>up</stem><lyric number="1" default-x="6.5" default-y="-48.69" relative-y="-30"><syllabic>begin</syllabic><text>glow</text></lyric></note><note default-x=
"206.68" default-y="-30"><pitch><step>G</step><octave>4</octave></pitch><duration>2</duration><voice>1</voice><type>eighth</type><stem>up</stem><lyric number="1" default-x="6.5" default-y="-48.69" relative-y="-30"><syllabic>end</syllabic><text>worm</text></lyric></note></measure><measure number="7" width="260.24"><note default-x="10.00" default-y="-40"><pitch><step>E</step><octave>4</octave></pitch><duration>4</duration><voice>1</voice><type>quarter</type><stem>up</stem><lyric number="1" default
-x="6.5" default-y="-48.69" relative-y="-30"><syllabic>single</syllabic><text>lend</text></lyric></note><note default-x="134.22" default-y="-40"><pitch><step>E</step><octave>4</octave></pitch><duration>2</duration><voice>1</voice><type>eighth</type><stem>up</stem><lyric number="1" default-x="8.81" default-y="-48.69" relative-y="-30"><syllabic>single</syllabic><text>thee,</text></lyric></note><note default-x="196.33" default-y="-40"><pitch><step>E</step><octave>4</octave></pitch><duration>2</dura
tion><voice>1</voice><type>eighth</type><stem>up</stem><lyric number="1" default-x="6.5" default-y="-48.69" relative-y="-30"><syllabic>single</syllabic><text>The</text></lyric></note></measure><measure number="8" width="270.59"><note default-x="10.00" default-y="-25"><pitch><step>A</step><octave>4</octave></pitch><duration>3</duration><voice>1</voice><type>eighth</type><dot default-x="34.55" default-y="-25" /><stem>up</stem><lyric number="1" default-x="6.5" default-y="-48.69" relative-y="-30"><s
yllabic>begin</syllabic><text>shoot</text></lyric></note><note default-x="103.17" default-y="-20"><pitch><step>B</step><octave>4</octave></pitch><duration>1</duration><voice>1</voice><type>16th</type><stem>up</stem><lyric number="1" default-x="6.5" default-y="-48.69" relative-y="-30"><syllabic>end</syllabic><text>ing</text></lyric></note><note default-x="144.57" default-y="-25"><pitch><step>A</step><octave>4</octave></pitch><duration>2</duration><voice>1</voice><type>eighth</type><stem>up</stem>
<lyric number="1" default-x="6.5" default-y="-48.69" relative-y="-30"><syllabic>single</syllabic><text>stars</text></lyric></note><note default-x="206.68" default-y="-30"><pitch><step>G</step><octave>4</octave></pitch><duration>2</duration><voice>1</voice><type>eighth</type><stem>up</stem><lyric number="1" default-x="6.5" default-y="-48.69" relative-y="-30"><syllabic>begin</syllabic><text>at</text></lyric></note></measure><measure number="9" width="323.99"><print new-system="yes"><system-layout>
<system-margins><left-margin>0</left-margin><right-margin>0</right-margin></system-margins><system-distance>125.05</system-distance></system-layout></print><note default-x="61.36" default-y="-40"><pitch><step>E</step><octave>4</octave></pitch><duration>4</duration><voice>1</voice><type>quarter</type><stem>up</stem><lyric number="1" default-x="6.5" default-y="-47.63" relative-y="-30"><syllabic>end</syllabic><text>tend</text></lyric></note><note default-x="181.74" default-y="-40"><pitch><step>E</s
tep><octave>4</octave></pitch><duration>2</duration><voice>1</voice><type>eighth</type><stem>up</stem><lyric number="1" default-x="9.02" default-y="-47.63" relative-y="-30"><syllabic>single</syllabic><text>thee;</text></lyric></note><note default-x="241.93" default-y="-40"><pitch><step>E</step><octave>4</octave></pitch><duration>1</duration><voice>1</voice><type>16th</type><stem>up</stem><lyric number="1" default-x="6.5" default-y="-47.63" relative-y="-30"><syllabic>single</syllabic><text>And</t
ext></lyric></note><note default-x="282.06" default-y="-40"><pitch><step>E</step><octave>4</octave></pitch><duration>1</duration><voice>1</voice><type>16th</type><stem>up</stem><lyric number="1" default-x="6.5" default-y="-47.63" relative-y="-30"><syllabic>single</syllabic><text>the</text></lyric></note></measure><measure number="10" width="264.46"><note default-x="21.89" default-y="-45"><pitch><step>D</step><octave>4</octave></pitch><duration>2</duration><voice>1</voice><type>eighth</type><stem
>up</stem><lyric number="1" default-x="6.5" default-y="-47.63" relative-y="-30"><syllabic>single</syllabic><text>elves</text></lyric></note><note default-x="82.09" default-y="-45"><pitch><step>D</step><octave>4</octave></pitch><duration>2</duration><voice>1</voice><type>eighth</type><stem>up</stem><lyric number="1" default-x="6.5" default-y="-47.63" relative-y="-30"><syllabic>begin</syllabic><text>al</text></lyric></note><note default-x="142.28" default-y="-30"><pitch><step>G</step><octave>4</oc
tave></pitch><duration>2</duration><voice>1</voice><type>eighth</type><stem>up</stem><lyric number="1" default-x="9.02" default-y="-47.63" relative-y="-30"><syllabic>end</syllabic><text>so,</text></lyric></note><note default-x="202.47" default-y="-30"><pitch><step>G</step><octave>4</octave></pitch><duration>2</duration><voice>1</voice><type>eighth</type><stem>up</stem><lyric number="1" default-x="6.5" default-y="-47.63" relative-y="-30"><syllabic>single</syllabic><text>Whose</text></lyric></note
></measure><measure number="11" width="292.7"><note default-x="10.00" default-y="-45"><pitch><step>D</step><octave>4</octave></pitch><duration>1</duration><voice>1</voice><type>16th</type><stem>up</stem><lyric number="1" default-x="6.5" default-y="-47.63" relative-y="-30"><syllabic>begin</syllabic><text>lit</text></lyric></note><note default-x="50.13" default-y="-40"><pitch><step>E</step><octave>4</octave></pitch><duration>1</duration><voice>1</voice><type>16th</type><stem>up</stem><lyric number
="1" default-x="6.5" default-y="-47.63" relative-y="-30"><syllabic>end</syllabic><text>tle</text></lyric></note><note default-x="90.26" default-y="-45"><pitch><step>D</step><octave>4</octave></pitch><duration>2</duration><voice>1</voice><type>eighth</type><stem>up</stem><lyric number="1" default-x="6.5" default-y="-47.63" relative-y="-30"><syllabic>single</syllabic><text>eyes</text></lyric></note><note default-x="150.45" default-y="-40"><pitch><step>E</step><octave>4</octave></pitch><duration>2<
/duration><voice>1</voice><type>eighth</type><stem>up</stem><lyric number="1" default-x="6.5" default-y="-47.63" relative-y="-30"><syllabic>single</syllabic><text>glow</text></lyric></note><note default-x="210.64" default-y="-30"><pitch><step>G</step><octave>4</octave></pitch><duration>1</duration><voice>1</voice><type>16th</type><stem>up</stem><lyric number="1" default-x="6.5" default-y="-47.63" relative-y="-30"><syllabic>single</syllabic><text>Like</text></lyric></note><note default-x="250.77"
 default-y="-30"><pitch><step>G</step><octave>4</octave></pitch><duration>1</duration><voice>1</voice><type>16th</type><stem>up</stem><lyric number="1" default-x="6.5" default-y="-47.63" relative-y="-30"><syllabic>single</syllabic><text>the</text></lyric></note></measure><measure number="12" width="252.57"><direction placement="above"><direction-type><wedge type="crescendo" default-y="25" spread="13" number="1" /></direction-type></direction><note default-x="10.00" default-y="-25"><pitch><step>A
</step><octave>4</octave></pitch><duration>2</duration><voice>1</voice><type>eighth</type><stem>up</stem><lyric number="1" default-x="6.5" default-y="-47.63" relative-y="-30"><syllabic>single</syllabic><text>sparks</text></lyric></note><note default-x="70.19" default-y="-15"><pitch><step>C</step><octave>5</octave></pitch><duration>2</duration><voice>1</voice><type>eighth</type><stem>down</stem><lyric number="1" default-x="6.5" default-y="-47.63" relative-y="-30"><syllabic>single</syllabic><text>
of</text></lyric></note><note default-x="130.39" default-y="-20"><pitch><step>B</step><octave>4</octave></pitch><duration>2</duration><voice>1</voice><type>eighth</type><stem>down</stem><lyric number="1" default-x="8.81" default-y="-47.63" relative-y="-30"><syllabic>single</syllabic><text>fire,</text></lyric></note><note default-x="190.58" default-y="-30"><pitch><step>G</step><octave>4</octave></pitch><duration>2</duration><voice>1</voice><type>eighth</type><stem>up</stem><lyric number="1" defau
lt-x="6.5" default-y="-47.63" relative-y="-30"><syllabic>single</syllabic><text>the</text></lyric></note><direction placement="above"><direction-type><wedge type="stop" number="1" /></direction-type></direction></measure><measure number="13" width="313.11"><print new-system="yes"><system-layout><system-margins><left-margin>0</left-margin><right-margin>0</right-margin></system-margins><system-distance>125.05</system-distance></system-layout></print><note default-x="61.36" default-y="-5"><pitch><s
tep>E</step><octave>5</octave></pitch><duration>3</duration><voice>1</voice><type>eighth</type><dot default-x="77.85" default-y="-5" /><stem>down</stem><beam number="1">begin</beam><notations><slur type="start" bezier-x="28.28489" bezier-y="16.528239" number="1" /></notations><lyric number="1" default-x="0.84" default-y="-48.4" relative-y="-30"><syllabic>single</syllabic><text>sparks,</text><extend /></lyric></note><note default-x="151.34" default-y="-10"><pitch><step>D</step><octave>5</octave><
/pitch><duration>1</duration><voice>1</voice><type>16th</type><stem>down</stem><beam number="1">end</beam><beam number="2">backward hook</beam></note><note default-x="191.33" default-y="-15"><pitch><step>C</step><octave>5</octave></pitch><duration>2</duration><voice>1</voice><type>eighth</type><stem>down</stem><notations><slur type="stop" number="1" bezier-x="-25.42371" bezier-y="20.660413" /></notations></note><note default-x="251.32" default-y="-30"><pitch><step>G</step><octave>4</octave></pit
ch><duration>2</duration><voice>1</voice><type>eighth</type><stem>up</stem><lyric number="1" default-x="6.5" default-y="-48.4" relative-y="-30"><syllabic>single</syllabic><text>the</text></lyric></note></measure><measure number="14" width="273.79"><note default-x="12.04" default-y="-25"><pitch><step>A</step><octave>4</octave></pitch><duration>3</duration><voice>1</voice><type>eighth</type><dot default-x="28.53" default-y="-25" /><stem>up</stem><beam number="1">begin</beam><notations><slur type="
start" orientation="over" placement="above" bezier-x="107.442625" bezier-y="38.184258" number="1" /></notations><lyric number="1" default-y="-48.4" relative-y="-30"><syllabic>single</syllabic><text>sparks</text><extend /></lyric></note><note default-x="102.02" default-y="-20"><pitch><step>B</step><octave>4</octave></pitch><duration>1</duration><voice>1</voice><type>16th</type><stem>up</stem><beam number="1">end</beam><beam number="2">backward hook</beam></note><note default-x="142.01" default-y=
"-25"><pitch><step>A</step><octave>4</octave></pitch><duration>3</duration><voice>1</voice><type>eighth</type><dot default-x="158.51" default-y="-25" /><stem>up</stem><beam number="1">begin</beam></note><note default-x="232.00" default-y="-20"><pitch><step>B</step><octave>4</octave></pitch><duration>1</duration><voice>1</voice><type>16th</type><stem>up</stem><beam number="1">end</beam><beam number="2">backward hook</beam></note></measure><measure number="15" width="275.07"><note default-x="13.32
" default-y="-30"><pitch><step>G</step><octave>4</octave></pitch><duration>3</duration><voice>1</voice><type>eighth</type><dot default-x="29.81" default-y="-25" /><stem>up</stem><beam number="1">begin</beam></note><note default-x="103.30" default-y="-25"><pitch><step>A</step><octave>4</octave></pitch><duration>1</duration><voice>1</voice><type>16th</type><stem>up</stem><beam number="1">end</beam><beam number="2">backward hook</beam></note><note default-x="143.30" default-y="-35"><pitch><step>F</
step><octave>4</octave></pitch><duration>3</duration><voice>1</voice><type>eighth</type><dot default-x="159.79" default-y="-35" /><stem>up</stem><beam number="1">begin</beam></note><note default-x="233.28" default-y="-30"><pitch><step>G</step><octave>4</octave></pitch><duration>1</duration><voice>1</voice><type>16th</type><stem>up</stem><beam number="1">end</beam><beam number="2">backward hook</beam></note></measure><measure number="16" width="271.75"><note default-x="10.00" default-y="-40"><pit
ch><step>E</step><octave>4</octave></pitch><duration>4</duration><voice>1</voice><type>quarter</type><stem>up</stem><notations><slur type="start" bezier-x="28.229942" bezier-y="-20.35162" number="2" /></notations></note><note default-x="149.97" default-y="-45"><pitch><step>D</step><octave>4</octave></pitch><duration>2</duration><voice>1</voice><type>eighth</type><stem>up</stem><notations><slur type="stop" number="1" bezier-x="-103.85417" bezier-y="47.077243" /><slur type="stop" number="2" bezier
-x="-29.610089" bezier-y="-18.285533" /></notations></note><note default-x="209.96" default-y="-25"><pitch><step>A</step><octave>4</octave></pitch><duration>2</duration><voice>1</voice><type>eighth</type><stem>up</stem><lyric number="1" default-x="6.5" default-y="-48.4" relative-y="-30"><syllabic>single</syllabic><text>of</text></lyric></note></measure></part></score-partwise>
\end{lstlisting}

  \bibliographystyle{ieeenat_fullname}
  \bibliography{main}
\else
  \maketitle

\begin{abstract}
Reconstructing an editable music score requires musical events, sung text and the instructions that organise them on the page. Existing recognition targets often omit part of this information, while direct MusicXML prediction repeats substantial structural markup. We introduce Cecilia, a compact representation of canonicalised MusicXML that retains musical content, lyric attachments and layout instructions. Cecilia-OMR predicts this representation through metadata and measure-bounded music segments from shared visual memory. A learned count head determines the number of output segments. Across paired complete scores under a common tokeniser, the ratio of mean MusicXML to Cecilia sequence lengths is 27.73. Cecilia-OMR outperforms Legato 2 on Camera String Quartets and both Lieder conditions, with a four-condition OpenScore macro normalised edit distance of 0.2232 against 0.2998. Separate evaluations of lyrics, notation trees and rendered layout reveal complementary aspects of reconstruction and quantify the contribution of layout instructions. Training-stage, reward and candidate-selection studies further examine how recognition changes across model adaptation and decoding.
\end{abstract}

\begin{figure}[t]
\centering
\includegraphics[width=\columnwidth]{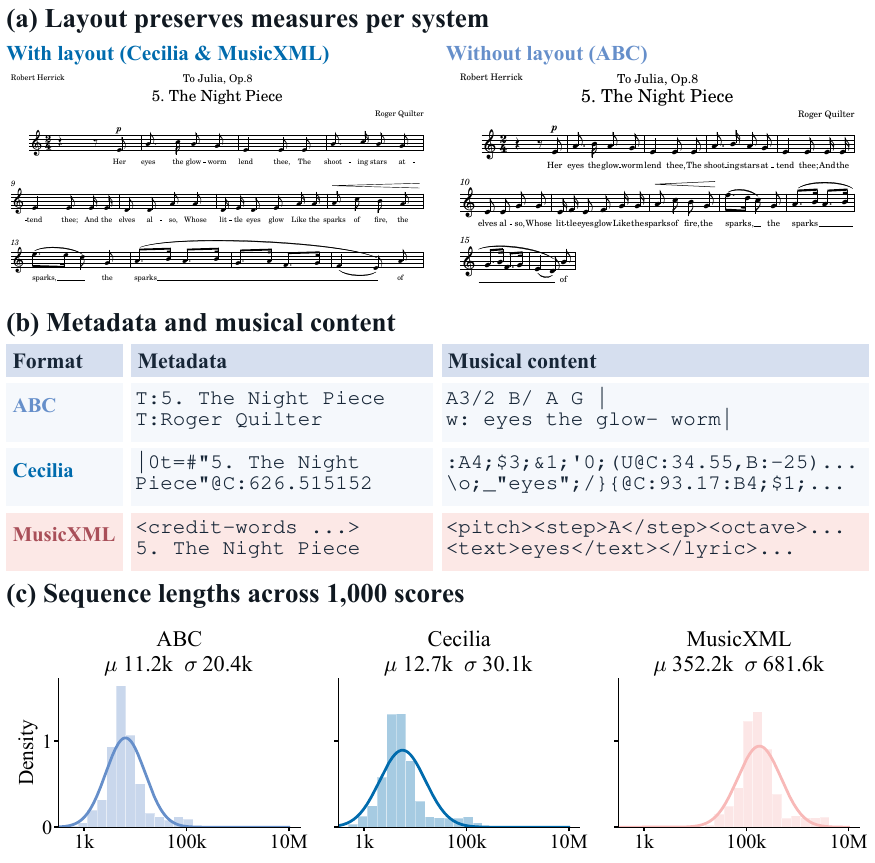}
\caption{\textbf{Differences across three formats.} (a) Layout. (b) Content. (c) Token lengths; $\mu$ and $\sigma$ denote the mean and standard deviation of token counts.}
\label{fig:representation}
\vspace{7pt}
\end{figure}

\begin{figure*}[t]
\centering
\includegraphics[width=\textwidth]{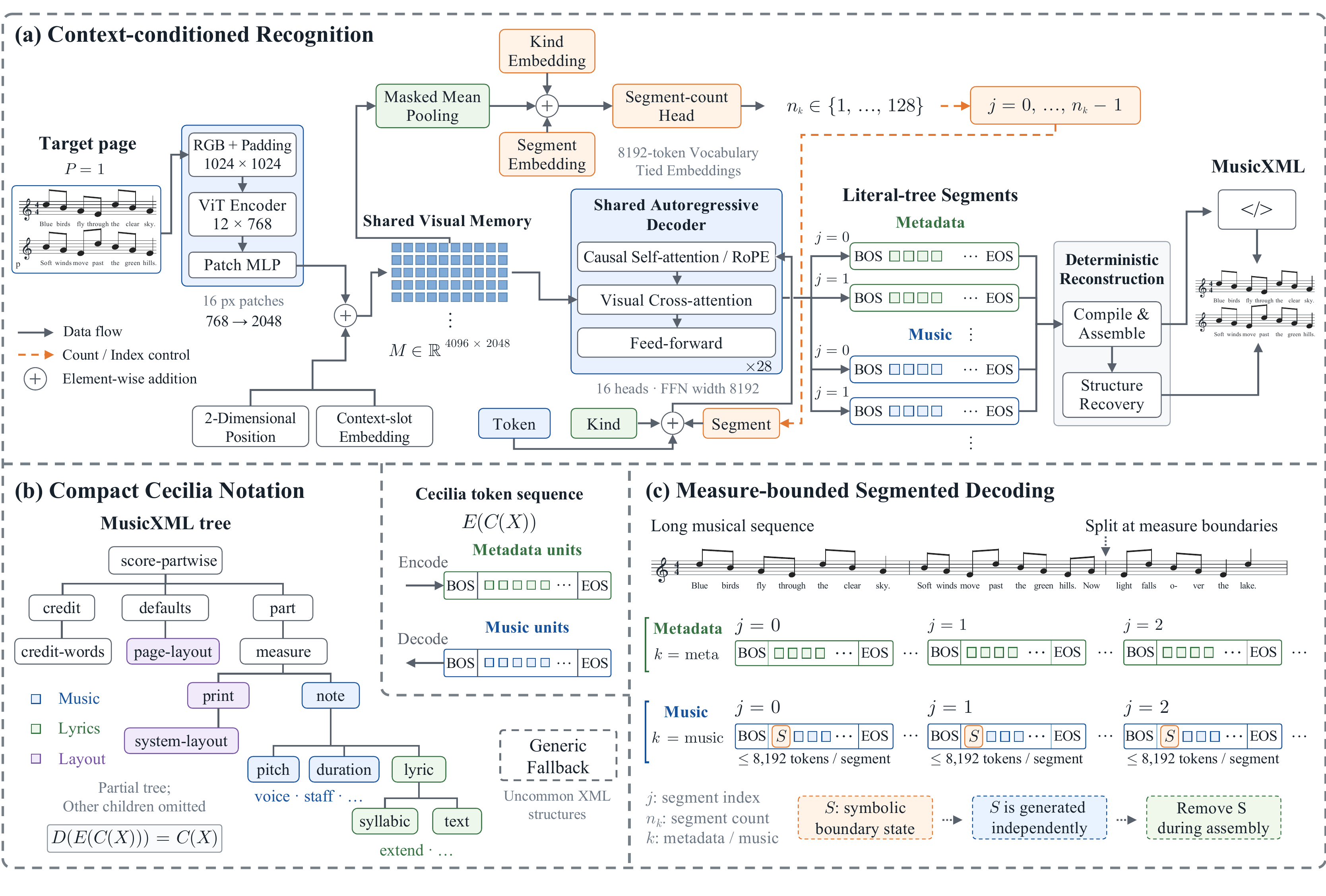}
\caption{\textbf{Cecilia-OMR shares visual memory across structured output segments.} (a) An encoder maps the input image to shared visual memory. Kind and segment embeddings condition the decoder, while a count head predicts segment counts. (b) Compact Cecilia notation reconstructs the retained MusicXML tree. (c) Music segments follow measure boundaries and independently predict boundary states, reusing visual memory with separate self-attention caches.}
\label{fig:architecture}
\end{figure*}

\vspace{7pt}
\section{Introduction}
\label{sec:intro}
An editable score encodes both a performance and a way of reading it. Lyric syllables attach to notes, measures form systems, and page breaks organise a work for the musician. End-to-end optical music recognition (OMR) translates score images into symbolic sequences: Sheet Music Transformer and Legato 1 establish full-page transcription, while Legato 2 extends recognition across diverse score images~\cite{riosvila2024sheet,yang2026legato,yang2026legato2}. Reconstructing the score as a musical document requires its content and written organisation to survive that translation.

The recognition target determines which of these details survive. ABC supports musical events, lyrics and formatting directives~\cite{walshaw2011abc}, but a rendering program must supply defaults for source attributes omitted by a transcription. Legato 1 replaces textual annotations with placeholders, while Legato 2 excludes lyrics from its recognition data~\cite{yang2026legato,yang2026legato2}. MusicXML offers explicit notation and layout fields~\cite{w3c2021musicxml40}: stem direction, notehead appearance, measure width and system or page breaks can accompany the notes. Predicting the complete markup directly repeats element and attribute names throughout a score. Richer targets therefore increase the sequence length precisely where polyphony, text and page organisation already make recognition demanding.

Repeated score structure offers a compact way to preserve these details. We introduce Cecilia (\textbf{C}ontext-\textbf{E}nhanced \textbf{C}onversion of \textbf{I}mages into \textbf{L}ayout-\textbf{I}nclusive Musical \textbf{A}nnotations), a compact notation for OMR. It assigns compact codes to recurring structures in the retained canonical MusicXML tree, including musical events, lyric attachments and layout instructions. Uncommon structures use a generic fallback, so the notation can retain content outside its specialised vocabulary. Figure~\ref{fig:representation} illustrates this balance: Cecilia retains musical content and layout instructions at a sequence length close to ABC. Across 1,000 paired full scores, the ratio of mean MusicXML to Cecilia sequence lengths is 27.73 under the same tokeniser. This reduction comes from the representation of structure, while musical values, text and layout remain in the target. The same passage in all three formats appears in the supplementary material (SM), Section~\ref{SM-sec:format_example}.

We pair this notation with Cecilia-OMR, a recognition model with approximately 2 billion parameters. Shared visual features condition measure-bounded segments that accommodate dense musical content. A shared encoder and decoder generate separate metadata and music units, conditioned by unit kind and segment index. Each segment reuses the unit's visual memory with its own text context, and a count head predicts how many segments the unit requires. This design connects shared visual evidence to manageable prediction sequences. It also recovers the lyric and layout fields that distinguish our target from Legato 1's text placeholders and Legato 2's lyric-free transcription.

The resulting model outperforms Legato 2 results on three OpenScore conditions and supports a separate assessment of page reconstruction (Table~\ref{tab:main}). Its four-domain macro normalised edit distance for OMR (OMR-NED) is 0.2232, compared with 0.2998 from Legato 2's domain scores. The improvements cover Camera String Quartets and both Lieder conditions, while our extended evaluation also includes vocal staves with lyrics and expert-transcribed handwriting. Break agreement and object placement compare the final layout, revealing whether a recognised sequence also retains the information needed to reconstruct a readable score.

Layout-aware OMR extends note transcription by requiring the joint recovery of musical content and its spatial organisation. Our contributions are fourfold: 1) Cecilia, a compact notation that jointly represents musical events and layout instructions while enabling reconstruction of the retained MusicXML tree; 2) Cecilia-OMR, a recogniser that combines shared visual memory, learned segment counts and measure-bounded generation; 3) An evaluation of notation accuracy and layout reconstruction; 4) Expert-transcribed MusicXML files for CVC-MUSCIMA, which we will release to support handwritten OMR research.

\section{Related Work}
\paragraph{Image-to-sequence music recognition.}
Early neural OMR combines convolutional visual encoders with sequence objectives for staff-level recognition~\cite{calvozaragoza2018primus,vanderwel2017optical}. Later methods address polyphony and full-page structure through autoregressive recognition~\cite{riosvila2023grandstaff,riosvila2024sheet,riosvila2026endtoend}. Legato 1 studies large-scale generalisable typeset OMR, and Legato 2 broadens its recognition setting~\cite{yang2026legato,yang2026legato2}. Cecilia-OMR shares the image-to-sequence formulation but makes the retained MusicXML tree the prediction target, including lyrics and layout attributes.

\paragraph{Musical representations and structured layout.}
MusicXML, MEI and ABC support different interchange and notation requirements~\cite{good2001musicxml,roland2002mei,walshaw2011abc}; linearised MusicXML also provides a target for end-to-end OMR~\cite{mayer2024practical}. Models such as Donut, Pix2Struct and Nougat demonstrate the value of structured textual outputs from images~\cite{kim2022donut,lee2023pix2struct,blecher2024nougat}. Our representation exploits repeated score-tree structures while retaining a fallback for uncommon XML content. The representation encodes the musical, textual and layout fields used for recognition.

\paragraph{Evaluation and annotated scores.}
OpenScore Lieder and String Quartets provide paired symbolic scores and renderable sources~\cite{gotham2022openscore,gotham2023quartets}. CVC-MUSCIMA provides handwritten score images for OMR~\cite{fornes2012cvcmuscima}. Symbolic edit metrics and notation-aware comparisons quantify different aspects of recognition~\cite{byrd2015testbed,foscarin2019diff}. We complement these resources with expert-transcribed MusicXML for CVC-MUSCIMA and explicit comparisons of final layout geometry.

\section{Methodology}
\label{sec:method}
Cecilia-OMR reconstructs an editable score from an input image (Figure~\ref{fig:architecture}). The Cecilia representation defines which musical, textual and layout fields to recover. Shared visual features condition metadata and measure-bounded music segments. Assembly combines the predicted metadata and musical content to produce the score.

\subsection{A compact canonical-tree representation}
We encode notation and layout through the repeated tree structures of MusicXML. Let $C(X)$ denote the retained canonical tree of a MusicXML document $X$. The Cecilia encoder $E$ and decoder $D$ reconstruct the encoded tree. For codec-admissible trees whose coordinates are exactly represented by the layout encoding,
\begin{equation}
D(E(C(X)))=C(X).
\end{equation}
Canonicalisation removes source-provenance and selected MIDI playback fields while retaining notes, lyrics, titles, credits and layout, including non-printing elements. Instrument identifiers and XML formatting are normalised. Horizontal differences use six-decimal precision. All canonical trees in the length study are reconstructed exactly, including attribute values, text and child order (SM Section~\ref{SM-sec:codec_roundtrip}). Algorithm~\ref{alg:codec} uses fixed structural codes, inline values and shared layout state for both directions.

Cecilia uses fixed codes for common elements and attributes, specialised containers for measures, notes, directions and lyrics, and literal values where required. A generic parenthesised form represents uncommon structures. Horizontal note positions are stored as differences from the preceding note on the same staff, with the position state reset at part and system boundaries. Vertical positions are derived from a recorded staff baseline and diatonic pitch only when the derived value exactly matches the original; otherwise they remain explicit. Lyrics remain attached to their note elements together with their syllabic and extension attributes, preserving both the text and its musical alignment. The resulting sequence keeps musical events, lyric attachments and layout attributes together in their authored tree structure.

A lyric syllable is encoded as a child of its note, and a system break at its authored measure. These associations are tree properties shared by the compact code and reconstructed XML. The generic branch uses the same ordered traversal for structures outside the specialised vocabulary.

The retained fields connect recognition to score rendering~\cite{good2001musicxml}. A \texttt{stem} element specifies direction, while \texttt{notehead} can express an explicit shape or fill choice; ordinary filled or hollow heads can also follow from note duration. The \texttt{width} attribute of a measure expresses its rendered width in tenths, and \texttt{print} attributes \texttt{new-system} and \texttt{new-page} specify breaks. System margins and distances describe the surrounding spacing. These instructions distinguish two scores with the same note sequence but different written presentation. The renderer uses these fields as explicit constraints when reconstructing the page.

Cecilia-OMR predicts this literal-tree sequence with lexical values written inline. The reader reconstructs the canonical tree, which is then exported as MusicXML. Fixed structural codes replace recurring element and attribute names; strings remain within their local tree context. The model does not predict per-score string dictionaries or repeated-record tables.

\begin{algorithm}[t]
\caption{MusicXML--Cecilia literal-tree conversion}
\label{alg:codec}
\begin{algorithmic}[1]
\Function{XML2Cecilia}{$X$}
  \State $T \gets C(X)$ \Comment{Retained canonical tree}
  \State $b \gets \Call{EncodeLiteralTree}{T,\ell_0}$
  \State \Return $\Call{VersionHeader}{}\,\Vert\,b$
\EndFunction
\Function{Cecilia2XML}{$c$}
  \State $b \gets \Call{ReadVersionHeader}{c}$
  \State $(T,r) \gets \Call{DecodeLiteralTree}{b,\ell_0}$
  \State \textbf{require} $r=\varnothing$ \Comment{No trailing input}
  \State \Return $\Call{SerializeXML}{T}$
\EndFunction
\end{algorithmic}
\small Fixed codes identify common structures; strings and other lexical values occur inline. The shared layout state $\ell_0$ reconstructs positions. Traversal preserves child order, and uncommon structures use the generic tree form.
\end{algorithm}

\subsection{Metadata and measure-bounded music units}
Compression reduces markup, while segmentation accommodates the remaining musical sequence. We divide music at authored measure boundaries. Each segment includes a predicted boundary state specifying the divisions value needed to interpret durations in its opening measure. The state is carried separately from the authored measures and is applied when the segment is exported independently. Assembly joins the musical content in segment order and removes the auxiliary boundary carriers. Metadata forms a separate unit with its own segment count. Each segment includes beginning and end tokens and fits within 8,192 tokens.

Metadata and music use the same visual encoder and decoder, with unit-kind embeddings distinguishing their prediction targets.

\subsection{Visual memory and segmented decoding}
Each unit's segments reuse the same visual evidence. A ViT encodes the input image after aspect-ratio-preserving padding to $1024\times1024$. A projector maps patch features to decoder width, and two-dimensional positional embeddings preserve spatial location. For an input image $I$, the encoder and projector produce
\begin{equation}
M_I\in\mathbb{R}^{4096\times2048}.
\end{equation}
The autoregressive decoder combines causal self-attention with cross-attention to this memory. Given token $y_t$, segment index $j$ and unit kind $k$, its input is
\begin{equation}
h_t=e_{\mathrm{tok}}(y_t)+e_{\mathrm{seg}}(j)+e_{\mathrm{kind}}(k).
\end{equation}
A segment-count head receives the mean valid visual memory, the unit-kind embedding and the segment-zero embedding. It predicts between one and 128 segments. Network dimensions and parameter counts are given in SM Table~\ref{SM-tab:architecture}.

Metadata and music are separate recognition units, identified by $k$. For either kind, let $Y_k$ contain the predicted count $n_k$ and segments $Y_{k,j}=(y_{k,j,1},\ldots,y_{k,j,T_{k,j}})$. The shared model generates that unit as
\begin{equation}
\begin{split}
p(Y_k\mid I,k)={}&p(n_k\mid I,k)\\
&\prod_{j=0}^{n_k-1}\prod_{t=1}^{T_{k,j}}
p(y_{k,j,t}\mid y_{k,j,<t},M_I,k,j).
\end{split}
\end{equation}
Each requested segment starts at BOS with an independent causal text cache. Its kind and segment-index embeddings specify the output, while cross-attention reuses $M_I$. Metadata and music invoke this unit-level procedure separately with shared weights. The selected outputs are subsequently assembled into a score.

The count head specifies the extent of a unit before its segments are decoded. Increasing that count adds decoder sequences over the same visual features, while segment indices determine their assembly order. Boundary state makes each music segment interpretable with its own text context.

\begin{figure*}[!t]
\centering
\includegraphics[width=\textwidth]{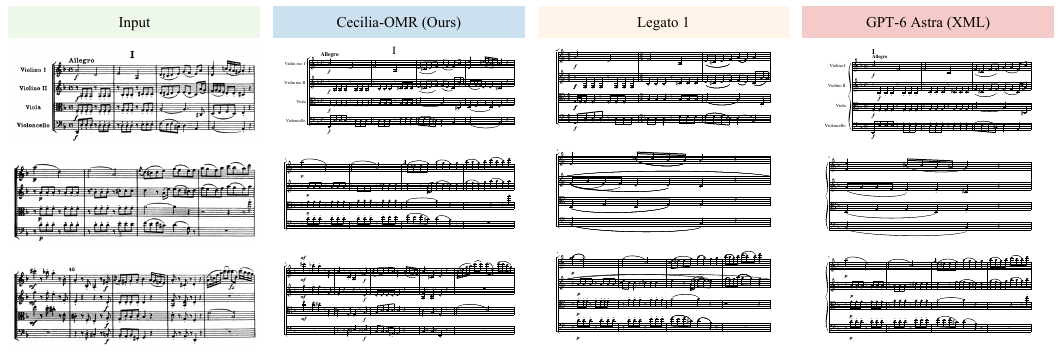}
\caption{\textbf{Score rendering makes layout differences visible.} We present the rendered first-page inference results of three models. The corresponding OMR-NED scores are 0.0188, 0.4072, and 0.0682 for Cecilia-OMR, Legato 1, and GPT-6 Astra (XML), respectively. Cecilia-OMR adopts a two-stage training strategy consisting of SFT followed by GRPO.}
\label{fig:qualitative}
\end{figure*}

\begin{table*}[t]
\centering\fontsize{8.5}{10}\selectfont
\setlength{\tabcolsep}{2pt}
\renewcommand{\arraystretch}{1.14}
\begin{tabular*}{\textwidth}{@{\extracolsep{\fill}}l*{7}{c}@{}}
\toprule
\multirow{3}{*}[-0.5ex]{Model} & \multicolumn{3}{c}{\textbf{Rendered}} & \multicolumn{3}{c}{\textbf{Camera}} & \textbf{Handwritten}\\
\cmidrule(lr){2-4}\cmidrule(lr){5-7}\cmidrule(l){8-8}
 & String Quartets & Lieder & Lieder with lyrics & String Quartets & Lieder & Lieder with lyrics & CVC-MUSCIMA\\
 & $N=252$ & $N=64$ & $N=64$ & $N=252$ & $N=64$ & $N=64$ & $N=100$\\
\midrule
GPT-6 Astra & 0.4468 & 0.3195 & 0.4486 & 0.4352 & 0.3548 & 0.4752 & 0.4061\\
GPT-6 Astra (XML) & 0.4131 & 0.3233 & 0.2430 & 0.4650 & 0.3698 & \textbf{0.2745} & \textbf{0.3297}\\
Claude Fable 5.1 & 0.8210 & 0.7666 & 0.6543 & 0.8453 & 0.8195 & 0.6983 & 0.7991\\
Gemini 4 Argon & 0.7265 & 0.6176 & 0.5613 & 0.7039 & 0.6238 & 0.5808 & 0.5512\\
Legato 1 & 0.3307 & 0.3554 & 0.4628 & 0.5835 & 0.5046 & 0.5820 & 0.8937\\
Legato 2 & \textbf{0.1710} & 0.2760 & \miss & 0.3160 & 0.4360 & \miss & \miss\\
Cecilia-OMR (Ours) & 0.2513 & \textbf{0.1839} & \textbf{0.2048} & \textbf{0.2307} & \textbf{0.2270} & 0.2862 & 0.5257\\
\bottomrule
\end{tabular*}
\caption{\textbf{Recognition OMR-NED across seven fixed subsets.} String Quartets, Lieder and Lieder with lyrics are OpenScore subsets. OMR-NED is a lower-is-better ratio. Bold marks the best score. A dash denotes an unavailable result. Cecilia-OMR uses SFT followed by GRPO.}
\label{tab:main}
\end{table*}

\begin{figure*}[t]
\centering
\includegraphics[width=\textwidth]{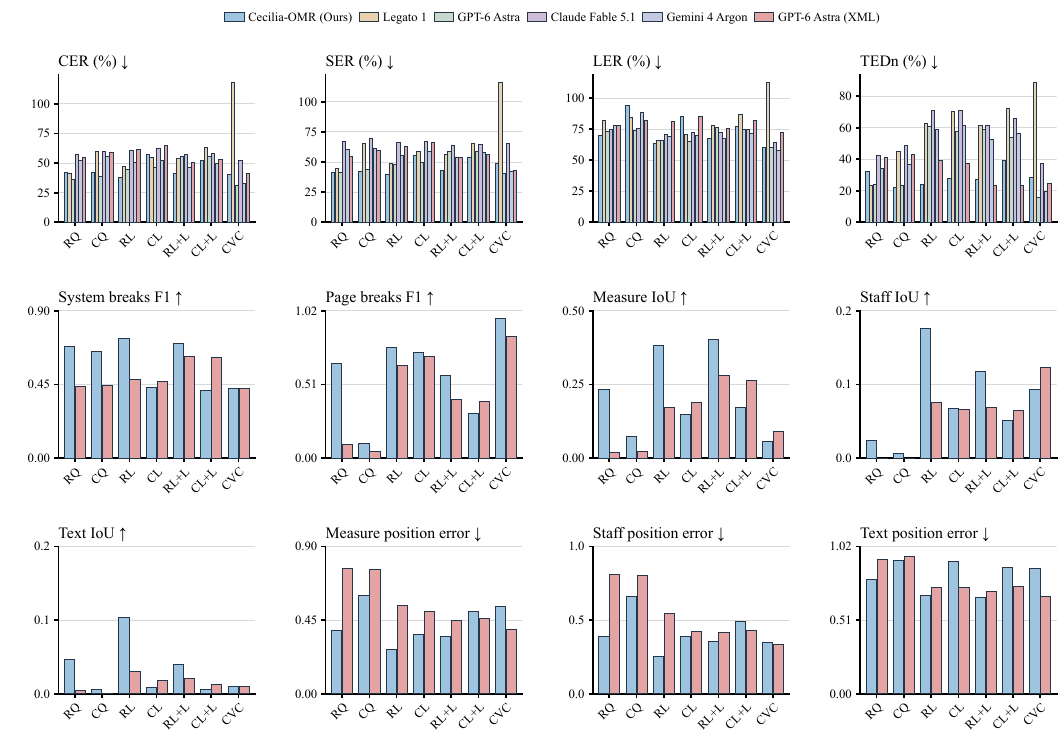}
\caption{\textbf{Recognition and layout measure complementary aspects of score reconstruction.} RQ: Rendered OpenScore String Quartets; CQ: Camera OpenScore String Quartets; RL: Rendered OpenScore Lieder; CL: Camera OpenScore Lieder; RL+L and CL+L: the corresponding conditions with lyrics; CVC: CVC-MUSCIMA. CER, SER, LER and TEDn are percentages. Arrows indicate the preferred direction. Layout metrics compare Cecilia-OMR after SFT and GRPO with direct MusicXML prediction by GPT-6 Astra.}
\label{fig:metrics}
\end{figure*}

\subsection{Optimisation and output assembly}
Learning must recover both the content of each segment and the extent of the output unit. Supervised training therefore combines next-token cross-entropy with segment-count classification and optimises the visual encoder, projector, decoder, conditioning embeddings and count head:
\begin{equation}
\mathcal{L}_{\mathrm{sup}}=
\frac{1}{BC}\sum_{b,t\in\mathcal{V}_b}\!\mathrm{CE}(p_{b,t},y_{b,t})
+\frac{\lambda_c}{B}\sum_{b=1}^{B}\mathrm{CE}(q_b,n_b-1),
\end{equation}
where $B$ is the number of units, $\mathcal{V}_b$ contains non-padding target positions, $n_b$ is the target segment count, and $C$ is a fixed token normaliser. Supervised training of Cecilia-OMR uses $C=1295.1014$ and $\lambda_c=0.1$. The fixed normaliser preserves the contribution of every valid token, so longer targets contribute more token terms.

The training sequence comprises pretraining on rendered symbolic scores, supervised adaptation to OpenScore and CVC-MUSCIMA, and group-relative policy optimisation (GRPO). We use the pretraining corpus of NotaGen~\cite{wang2025notagen}, comprising approximately 1.6 million musical pieces, and convert the scores into Cecilia notation. The corpus covers diverse musical genres and instrumentations. Our preprocessing retains lyric text and its alignment with musical notes. GRPO samples candidate segment counts and token sequences, compares their rewards within each group, and regularises the policy towards a frozen reference. The reward is
\begin{equation}
R=0.95\left(1-\frac{\sum_i E_i}{\sum_i(G_i+P_i)}\right)+0.05F,
\end{equation}
where $E_i$, $G_i$ and $P_i$ are notation edits, reference sizes and prediction sizes for the compiled output before recovery. The format term $F$ is the fraction of observed metadata and music units that complete generation, compile from the literal representation and export valid XML. The candidate is generated from images, and reference XML supplies its reward. Reward evaluation uses the native compiled candidate before deterministic recovery. Exact settings and the SFT parent are specified in the SM, Section~\ref{SM-sec:optimisation}.

At inference, grammar-prefix decoding checks whether each candidate token extends a valid prefix of the literal-tree syntax and chooses the highest-scoring admissible continuation. Unconstrained and grammar-prefix candidates are compiled to MusicXML and selected using native XML availability. The subsequent structural recovery reconciles the selected output by restoring declarations implied by predicted content, retaining usable music subtrees and assembling metadata with the ordered music. SM Section~\ref{SM-sec:recovery} specifies the syntax and recovery rules.

\section{Experimental Setup}
\label{sec:experiments}
\paragraph{Datasets.}
We follow the evaluation setup of Legato 1 and Legato 2, using Rendered and Camera OpenScore String Quartets (252 pages each) and Rendered and Camera OpenScore Lieder (64 pages each)~\cite{yang2026legato,gotham2023quartets,gotham2022openscore}. Here, Rendered denotes images rendered from symbolic scores, whereas Camera denotes images extracted from source scans. The original Lieder conditions retain the piano part and mask the vocal staff. We use their released reference XML unchanged. We additionally evaluate rendered and camera \emph{OpenScore Lieder with lyrics} conditions, each comprising 64 pages with source-derived references that retain vocal staves and lyrics. These extensions jointly evaluate the vocal staff and its lyrics, adding both to the original piano-only reference. For handwriting, we evaluate CVC-MUSCIMA~\cite{fornes2012cvcmuscima} with expert-transcribed MusicXML reference annotations. Training and test sets are separated by musical work: two of the 20 scores are held out, comprising 100 test pages.

OpenScore test works are excluded from pretraining, SFT and GRPO. SFT uses the remaining OpenScore works, with training and test sets separated at the musical-work level.

\paragraph{Baselines.}
We evaluate the released full-size Legato 1 model and retain the four published Legato 2 OMR-NED values~\cite{yang2026legato,yang2026legato2}. The other baselines also provide predictions for the additional metrics. GPT-6 Astra~\cite{openai2026astra}, Claude Fable 5.1~\cite{anthropic2026fable51} and Gemini 4 Argon~\cite{google2026argon} produce ABC that is converted to MusicXML. GPT-6 Astra (XML) directly predicts MusicXML and is the API comparator for layout.

\paragraph{Evaluation.}
For domain $d$, OMR-NED aggregates edit counts $E_i$ and notation sizes $G_i,P_i$ as
\begin{equation}
\mathrm{OMR\text{-}NED}_d=\frac{\sum_{i\in d} E_i}{\sum_{i\in d}(G_i+P_i)}.
\end{equation}
The four-condition OpenScore summary is the unweighted mean of its four domain scores. CER, SER and LER compare characters, lexical tokens and text lines after a fixed XML-to-ABC conversion. TEDn compares notation trees. These four error rates are percentages and can exceed 100. Full definitions appear in SM Section~\ref{SM-sec:metrics}.

\paragraph{Incomplete outputs.}
Every evaluation page remains in its fixed domain. For a metric that cannot consume a model's output, we score an empty prediction using that metric's reference units; successful metrics for the same page remain unchanged. Final API service failures also count as empty outputs and are reported separately from invalid model outputs. The empty output contributes reference-only deletions. Reference errors and scoring failures are tracked separately from model failures and corrected before aggregation across each evaluation domain.

\paragraph{Layout geometry.}
We render predictions and references with identical MuseScore settings and fonts. Although camera layout recovery can be ill-posed, OpenScore XML reproduces the source layout, supporting Camera layout evaluation (Figure~\ref{fig:qualitative}). We measure system/page break agreement, measure/staff/text-word IoU and normalised position error (NPE) against rendered references. Unmatched objects receive zero IoU and maximal NPE. Cecilia-OMR and GPT-6 Astra (XML) use the same evaluation settings.

\section{Experimental Results}
\label{sec:results}
The representation analysis measures sequence length under a common tokeniser. The recognition and layout experiments below evaluate Cecilia-OMR as a complete system.

\subsection{Music and lyric recognition}
Cecilia-OMR outperforms Legato 2 on Camera String Quartets and both official Lieder conditions (Table~\ref{tab:main}). OMR-NED decreases from 0.3160 to 0.2307 on Camera String Quartets, from 0.2760 to 0.1839 on Rendered Lieder, and from 0.4360 to 0.2270 on Camera Lieder. These gains reduce the four-domain macro from 0.2998 to 0.2232. On Rendered String Quartets, Legato 2 obtains 0.1710 and Cecilia-OMR obtains 0.2513, locating the strongest gains in camera quartets and Lieder.

Retaining vocal staves and lyrics changes the relative strengths of the systems. Cecilia-OMR has lower OMR-NED than direct-XML Astra on the rendered extension (0.2048 versus 0.2430), while Astra leads on camera pages (0.2745 versus 0.2862). Cecilia-OMR's lyric-specific CER is 21.03\% on rendered images and 50.61\% on camera images, distinct from the ABC-based CER in Figure~\ref{fig:metrics}. These lyric errors distinguish transcription of sung words from aggregate notation accuracy. On handwritten CVC-MUSCIMA, Astra XML leads with 0.3297 OMR-NED against Cecilia-OMR's 0.5257.

\subsection{Decoding strategy}
Native XML availability provides a useful candidate-selection signal. With the same checkpoint, image windows and final output processing, selection between unconstrained and grammar-prefix candidates reduces OpenScore macro OMR-NED from 0.2591 to 0.2232 (Table~\ref{tab:decoding_strategy}). The macro with lyrics decreases from 0.3432 to 0.2455, and CVC-MUSCIMA OMR-NED from 0.5651 to 0.5257. The weights, input windows and final recovery are shared, so the comparison measures the effect of candidate selection.

\begin{table}[t]
\centering\small
\begin{tabular}{@{}lcc@{}}
\toprule
Evaluation & Unconstrained & XML selection\\
\midrule
OpenScore & 0.2591 & \textbf{0.2232}\\
Lieder with lyrics & 0.3432 & \textbf{0.2455}\\
CVC-MUSCIMA & 0.5651 & \textbf{0.5257}\\
\bottomrule
\end{tabular}
\caption{OMR-NED for two decoding strategies. OpenScore averages four domain ratios; Lieder with lyrics averages two; CVC-MUSCIMA uses one corpus ratio.}
\label{tab:decoding_strategy}
\end{table}

\subsection{Layout agreement}
Cecilia-OMR improves layout agreement over direct-XML Astra on rendered OpenScore conditions. On String Quartets, system-break F1 is 0.6829 versus 0.4404, and measure IoU is 0.2340 versus 0.0193. On Lieder, measure IoU is 0.3829 versus 0.1712. Break and placement metrics capture distinct properties: on CVC-MUSCIMA, page-break F1 reaches 0.9700 while measure IoU is 0.0556. On Lieder with lyrics, Cecilia-OMR has higher measure IoU on rendered pages (0.4017 versus 0.2809), while camera pages reverse this comparison (0.1737 versus 0.2623). The shared renderer turns the retained notation and layout fields into comparable pages. The gains therefore concern the reconstructed score presented to a reader, including its grouping of measures and their placement. A prediction can place a system break at the correct measure while distributing its staves or text differently within the page. The separate metrics distinguish correct measure grouping from accurate spatial placement. Figure~\ref{fig:qualitative} shows Cecilia-OMR following the input's measure grouping more closely; Legato 1 and Astra XML introduce an earlier system break.

Table~\ref{tab:layout_ablation} isolates the contribution of predicted layout instructions by rendering the same 48 predictions with and without these fields. Removing them reduces system-break F1 from 0.6917 to 0.2063 and measure IoU from 0.2734 to 0.1319. Predicted layout also improves text IoU and all three position errors, while automatic layout improves page-break F1 and staff IoU. Musical content and lyrics remain identical within each pair; implementation details appear in the SM.

\begin{table}[!ht]
\centering\footnotesize
\begin{tabular}{@{}lcc@{}}
\toprule
Metric & Predicted layout & Automatic layout\\
\midrule
System-break F1 $\uparrow$ & \textbf{0.6917} & 0.2063\\
Page-break F1 $\uparrow$ & 0.5417 & \textbf{0.6667}\\
Measure IoU $\uparrow$ & \textbf{0.2734} & 0.1319\\
Staff IoU $\uparrow$ & 0.0802 & \textbf{0.1043}\\
Text IoU $\uparrow$ & \textbf{0.0263} & 0.0120\\
Measure NPE $\downarrow$ & \textbf{0.3210} & 0.3641\\
Staff NPE $\downarrow$ & \textbf{0.3259} & 0.3500\\
Text NPE $\downarrow$ & \textbf{0.8271} & 0.8789\\
\bottomrule
\end{tabular}
\caption{Layout ablation on 48 pages with identical predicted music and lyrics. Only layout instructions differ.}
\label{tab:layout_ablation}
\end{table}

\subsection{Complementary recognition errors}
Cecilia-OMR's full-content TEDn is 24.32\% on rendered pages and 28.08\% on camera pages, compared with 39.45\% and 37.36\% for direct-XML Astra. The conditions with lyrics expose a different balance: Astra XML has lower TEDn on both, although Cecilia-OMR retains the lower rendered-condition OMR-NED. OMR-NED pools edits and notation counts across a domain, giving longer scores a larger contribution. TEDn normalises each page by the deletion cost of its reference tree before averaging. A model can therefore recover a larger share of notation overall while making more consequential tree errors on individual pages.

The lyric-preserving conditions add a sung line to the accompaniment. We therefore evaluate lyric characters from their XML attachments and rendered text positions separately, distinguishing transcription of the words from their placement on the page.

\begin{figure}[t]
\centering
\includegraphics[width=\columnwidth]{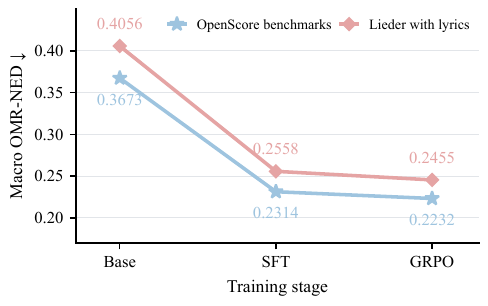}
\caption{\textbf{Training stages improve aggregate recognition in Cecilia-OMR.} Macro OMR-NED across Base, SFT and GRPO, averaging the four OpenScore benchmark conditions or the two Lieder conditions with lyrics.}
\label{fig:scaling_demo}
\end{figure}

\subsection{Training stages}
Supervised adaptation reduces OpenScore macro OMR-NED from 0.3673 to 0.2314, and GRPO further reduces it to 0.2232 (Figure~\ref{fig:scaling_demo}). The lyric-preserving macro follows the same progression, from 0.4056 to 0.2558 and 0.2455. SM Table~\ref{SM-tab:model_variants} gives the domain-level changes.
\ifarxivversion

Domain-level changes complement the macro averages. From SFT to GRPO, OMR-NED falls from 0.2187 to 0.1839 on Rendered Lieder and from 0.2336 to 0.2270 on Camera Lieder. Rendered Lieder with lyrics also improves, from 0.2383 to 0.2048, whereas the camera condition rises from 0.2732 to 0.2862. 
\fi

\subsection{Reward design}
Table~\ref{tab:reward_main} compares the SFT parent and three GRPO reward designs. Original uses $0.95M+0.05F$; Format increases the format weight to $0.90M+0.10F$. Lyric uses $0.80M+0.15L+0.05F$ on Lieder with lyrics, where $L=\max(0,1-E_\ell/G_\ell)$ measures lyric character accuracy, and retains the original reward on Quartets. Original gives the lowest OMR-NED on the four notation-only conditions, while Lyric gives the lowest error on both lyric-preserving conditions. The mean OMR-NED across the two lyric-preserving conditions falls from 0.2290 with Original to 0.1315 with Lyric and 0.1412 with Format.

\begin{table}[!ht]
\centering\footnotesize
\setlength{\tabcolsep}{2.5pt}
\begin{tabular*}{\columnwidth}{@{\extracolsep{\fill}}lcccc@{}}
\toprule
Subset & SFT & Original & Lyric & Format\\
\midrule
Rendered Quartets & 0.2418 & \textbf{0.2401} & 0.2657 & 0.2596\\
Camera Quartets & 0.2753 & \textbf{0.2533} & 0.3255 & 0.3089\\
Rendered Lieder & 0.3008 & \textbf{0.0841} & 0.2705 & 0.1263\\
Camera Lieder & 0.2643 & \textbf{0.2600} & 0.3909 & 0.3412\\
Rendered Lieder + lyrics & 0.1829 & 0.1880 & \textbf{0.1195} & 0.1301\\
Camera Lieder + lyrics & 0.1871 & 0.2701 & \textbf{0.1434} & 0.1524\\
\bottomrule
\end{tabular*}
\caption{Subset OMR-NED on eight pages per condition. SFT is the common initial model; Original, Lyric and Format use the respective reward designs. Bold marks the best score.}
\label{tab:reward_main}
\end{table}
\subsection{Sampling and candidate selection}
Increasing the candidate set from one to four lowers both macro errors at all three temperatures (Table~\ref{tab:sampling_main}). At $T=0.7$, four candidates yield OpenScore and lyric-preserving OMR-NED of 0.2696 and 0.2620. Intermediate changes depend on the condition: at $T=0.3$, two candidates give lower lyric-preserving error than four. At every candidate count, both lower temperatures outperform temperature 1.0.

The fixed ranking first prefers native XML validity and then ranks candidates by mean model log probability. Additional samples expand the available hypotheses, but the rule may prefer a different musical transcription as the set grows. This explains why candidate count and measured accuracy need not move together at every intermediate step.

\begin{table}[!ht]
\centering\small\setlength{\tabcolsep}{4pt}
\begin{tabular}{@{}lccc@{}}
\toprule
$T$ & $N=1$ & $N=2$ & $N=4$\\
\midrule
$0.3$ & 0.3212 / 0.3623 & 0.3133 / 0.2376 & 0.2822 / 0.2892\\
$0.7$ & 0.3405 / 0.4084 & 0.3220 / 0.3771 & 0.2696 / 0.2620\\
$1.0$ & 0.5406 / 0.7232 & 0.4263 / 0.7046 & 0.4281 / 0.5656\\
\bottomrule
\end{tabular}
\caption{Macro OMR-NED on 48 pages: OpenScore / lyric-preserving. Nested candidates pair metadata and music by seed; selection uses native validity and mean model log probability, with references used only for scoring.}
\label{tab:sampling_main}
\end{table}

\section{Conclusion and Discussion}
Cecilia opens a new direction for OMR through a compact notation that unifies musical content, lyric attachments and layout instructions. Cecilia-OMR predicts this representation with shared visual memory and segmented outputs, improving OMR-NED over Legato 2 on Camera String Quartets and both official Lieder conditions. Joint recognition and layout evaluation reveals how accurately an editable score preserves its content and presentation. Layout reconstruction still has substantial room for improvement. Future work will focus on more faithful page geometry, stronger lyric alignment and musical state prediction, and alternative reinforcement learning strategies for improving recognition and layout reconstruction. We will release expert-transcribed CVC-MUSCIMA MusicXML to support further research on handwritten music recognition.

\clearpage
\section*{LLM Usage Statement}
Large language models assist with manuscript drafting, language editing, code development and figure preparation. Responsibility for all scientific claims, experimental results and the final manuscript rests with the authors.

{\small
\bibliographystyle{ieeenat_fullname}
\bibliography{main}
}

  \ifarxivversion
    \clearpage
    \begingroup
    \setcounter{table}{0}
    \setcounter{figure}{0}
    \setcounter{equation}{0}
    \setcounter{algorithm}{0}
    \renewcommand{\theHsection}{supplement.\arabic{section}}
    \renewcommand{\theHtable}{supplement.\arabic{table}}
    \renewcommand{\theHfigure}{supplement.\arabic{figure}}
    \renewcommand{\theHequation}{supplement.\arabic{equation}}
    \let\CeciliaOriginalLabel\label
    \let\CeciliaOriginalRef\ref
    \renewcommand{\label}[1]{\CeciliaOriginalLabel{SM-#1}}
    \let\ref\CeciliaSupplementRef
    \twocolumn[{%
      \centering
      {\Large\bfseries Cecilia: Towards Layout-Inclusive Optical Music Recognition\par}
      \vspace{0.4em}
      {\large\bfseries Supplementary Material\par}
      \vspace{1.5em}
    }]
\setcounter{section}{0}
\renewcommand{\thesection}{\Alph{section}}
\renewcommand{\thetable}{S\arabic{table}}

\section{Token-length analysis}
We apply the same frozen Cecilia-OMR tokeniser to full-score ABC, literal-tree Cecilia and the corresponding MusicXML. Table~\ref{tab:tokens} reports population standard deviations ($\mathrm{ddof}=0$). The comparison contains 1,000 unique, successfully converted scores: 895 OpenScore Lieder and 105 String Quartets. All three formats must convert successfully for a score to enter the paired sample.

Token means for the complete score differ from the sum of split training-unit tokens because segmentation introduces headers, boundary state and special tokens.

\begin{table}[!htbp]
\centering\footnotesize
\begin{tabular}{@{}lrrr@{}}
\toprule
Format & Mean & Std. dev. & Median\\
\midrule
ABC & 11,175.86 & 20,413.60 & 5,626.5\\
Cecilia & 12,702.02 & 30,126.33 & 4,702.0\\
MusicXML & 352,228.32 & 681,560.15 & 156,222.0\\
\bottomrule
\end{tabular}
\caption{Token lengths for 1,000 paired complete scores under one fixed tokeniser. The statistics describe representation length under the common tokeniser.}
\label{tab:tokens}
\end{table}

The mean length of Cecilia is 1.14 times that of ABC while retaining the encoded score fields. The mean length of MusicXML is 27.73 times that of Cecilia. The token-length figure shows empirical histograms and descriptive Gaussian fits in log-ten token space. Its mean and standard-deviation annotations refer to raw token counts. The Gaussian curves summarise log-space shape, while raw lengths retain long right tails. A tokeniser trained separately for XML or ABC may change these ratios.

\section{Representation and target construction}
\subsection{Canonical-tree preservation}
Canonicalisation removes the \texttt{encoding}, \texttt{source}, \texttt{relation} and \texttt{miscellaneous} children of \texttt{identification}. Instrument identifiers are standardised consistently; device routing, MIDI channel, instrument-name playback overrides, volume, pan and elevation are removed. Musical notes, lyrics, title and credit text, instrument names, tempo, layout and \texttt{print-object="no"} elements remain. The structural pass normalises attribute order, line endings, formatting whitespace and negative-zero attributes. Algorithm~\ref{Main-alg:codec} decodes directly to the retained tree before any renderer-specific export transformations.

The decoder checks the literal-tree framing and complete consumption of the input, as specified in Algorithm~\ref{Main-alg:codec}. The generic form covers registered structures outside the specialised vocabulary. Horizontal differences use six-decimal rounding and reset at part and system boundaries. For example, positions 120, 145 and 170 are represented by 120, 25 and 25. A vertical baseline advances by five tenths per diatonic step; its coordinate is omitted only when the original value is reproduced exactly. An absence flag preserves a missing coordinate. Lyrics retain their explicit coordinates.

\subsection{Canonical-tree round-trip validation}
\label{sec:codec_roundtrip}
We test the literal-tree representation predicted by Cecilia-OMR on the same 1,000 complete scores used in the sequence-length comparison: 895 OpenScore Lieder and 105 String Quartets. Each canonical XML tree is encoded as a literal-tree sequence and reconstructed with its reader, without renderer-specific export transformations. The comparison includes element names, attribute keys and their string values, text, tails and ordered children. Canonicalisation sorts attributes, normalises line endings and negative-zero attributes, and removes whitespace-only formatting text.

All 1,000 reconstructed trees match their canonical inputs exactly, with zero exceptions and no numerical differences. This test evaluates the literal-tree encoder and reader directly, rather than only tokenisation or assembly of stored segments. The corpus contains 1,095,973 note horizontal-position attributes, all with two decimal places. Other numeric attributes have up to six decimal places.

The result is empirical for this corpus. Horizontal note differences are rounded to six decimal places. In a precision test, successive horizontal positions of 10.1234567 and 20.6543219 reconstruct as 10.1234567 and 20.6543219.

\subsection{Tokenisation and target admission}
Tokenisation uses frozen byte-level BPE with 8,192 vocabulary entries, no text normaliser and no regular-expression pre-tokeniser. The byte vocabulary encodes Unicode lyric text. A segment's 8,192-token budget includes beginning and end tokens. Training examples preserve complete authored measures. Boundary-state carriers contain the declarations needed for an independently decoded segment and are removed during assembly.

\section{Architecture and optimisation}
\subsection{Network and trainable parameters}
Table~\ref{tab:architecture} specifies the dimensions of the model in Section~\ref{Main-sec:method}. Decoder blocks use pre-normalisation and residual connections; text self-attention uses rotary positions. The visual encoder and projector retain the full patch grid.

\begin{table}[!htbp]
\centering\footnotesize
\begin{tabular}{@{}ll@{}}
\toprule
Component & Configuration\\
\midrule
Input & RGB, padded $1024\times1024$\\
Vision encoder & 12 blocks, width 768, patch 16\\
Memory per image & $4096\times2048$\\
Patch projector & $768\rightarrow2048\rightarrow2048$\\
Decoder & 28 blocks, width 2048\\
Attention & 16 heads, head width 128\\
Feed-forward width & 8192\\
Vocabulary & 8192; tied embedding/output\\
Segment conditions & 128 learned embeddings\\
Unit conditions & Metadata and music\\
Segment-count head & 128 classes\\
Maximum segment length & 8192 including BOS/EOS\\
\bottomrule
\end{tabular}
\caption{The active Cecilia-OMR architecture.}
\label{tab:architecture}
\end{table}

Supervised learning and policy updates optimise the visual encoder, projector, decoder, conditioning embeddings and segment-count head. GRPO uses a separately frozen copy as its reference policy.

Class $c$ of the segment-count head represents $c+1$ segments. The vocabulary projection uses FP32 computation, while the surrounding forward computation uses BF16. Cross-attention keys and values are cached for all segments of a unit.

\subsection{Training objectives and exposure}
\label{sec:optimisation}
The supervised objective in Section~\ref{Main-sec:method} uses the fixed token normaliser $C=1295.1014$ and count-classification weight 0.1. The following settings specify the subsequent policy refinement.

The GRPO stage starts from the mixed-data SFT model selected after 8,662 continuation updates. The policy and frozen reference initialise from identical SFT weights, and GRPO starts with fresh AdamW state. It samples four candidates for each work and uses two groups per optimiser batch across eight GPUs.

The sampled and replayed policies both use temperature 0.7. The optimiser uses learning rates $10^{-7}$ for non-visual parameters and $10^{-8}$ for the visual encoder, AdamW betas $(0.9,0.95)$, epsilon $10^{-8}$ and zero weight decay. The clipped policy ratio uses $\epsilon=0.2$ and the KL coefficient is 0.01.

Reward notation and format validity follow the definition in Section~\ref{Main-sec:method}. Format validity divides the number of valid metadata and music observation units by their fixed total. Unresolved scoring failures pause the affected optimisation group until scoring is available.

Within each four-sample group, advantages subtract the group mean and divide by the larger of the population reward standard deviation and 0.01. A standard deviation at most $10^{-8}$ skips the group entirely, including its KL term. The replay objective retains the fixed supervised token normaliser multiplied by the number of observed units, with fixed relative weights for metadata and music. 

The policy samples segment counts and tokens. Their replayed log probabilities define the group-relative update; reference notation supplies the reward. 

\subsection{Ground-truth-free decoding and recovery}
\label{sec:recovery}
Grammar-prefix decoding checks every proposed token against the running literal-tree reader state, including string encoding, numeric syntax and matching delimiters. The highest-scoring admissible continuation is selected. A greedy end-of-sequence decision is retained even when premature, and that candidate remains incomplete. Candidate selection uses native XML availability.

Recovery retains complete parsed subtrees and discards malformed trailing content. Metadata recovery combines compatible predicted roots and reconstructs part declarations from predicted identifiers. Music recovery uses divisions supplied by valid boundary carriers or compatible adjacent segments of the same part; duration-based inference is accepted only when the available constraints agree. These rules use generated content, and missing musical content remains absent. OMR-NED evaluates the resulting exported score.

The boundary carrier stores a positive divisions value for a contiguous run of measures from one part. Standalone export inserts this value into the first measure and removes the private carrier; full-score assembly removes the carrier while retaining the authored measures. It does not carry clefs, key signatures or time signatures.

\section{Evaluation}
\subsection{Reference scope}
The official OpenScore conditions retain their released reference XML~\cite{yang2026legato}. Canonicalisation for training does not modify these evaluation references. The additional Lieder references retain both the vocal staff and its lyrics.

CVC-MUSCIMA~\cite{fornes2012cvcmuscima} uses a work-disjoint split: two of its 20 musical scores are held out for testing, comprising 100 handwritten pages, while the remaining scores provide training data. Experts transcribe these scores into MusicXML, which we will release for further research.

\subsection{Baseline representations and inference}
The ABC and direct-XML baseline outputs described in the main paper are converted and rendered through the same downstream tools.

For the direct-XML condition, requests use the model identifier \texttt{gpt-6-astra} through \texttt{api.gpt.ge/v1/chat/completions}. Requests set temperature to zero and the output limit to 64,000 tokens. Each request contains one user message with the prompt and a PNG image.

Failed or incomplete requests are retried. Outputs are selected by the first completed response, without consulting reference scores. The following is the XML transcription prompt:

\begin{quote}\footnotesize
You are an optical music recognition system. The image is one page of printed sheet music.
Transcribe the full page into MusicXML.

Requirements:

Output one complete MusicXML score-partwise document, with a part-list and parts, including the printed title, meter, key and tempo, and valid divisions and note durations.

For a score with more than one staff, declare the corresponding parts, staves and voices in MusicXML, group them to match the printed bracing, and assign each note to its correct voice and staff.

Transcribe every staff, every system and every measure that is printed on this page, in score order. Do not invent music that is not printed, and do not skip printed music.

Include pitches, rests, durations, bar lines, clefs, key signatures, time signatures, accidentals, beams, ties, slurs, articulations, dynamics, tempo and other text directions.

If the page has sung text, write it in lyric elements attached to the notes of the voice that sings it. If the page has no sung text, do not add any lyric elements.

Output MusicXML only. No explanation, no commentary, no markdown code fences.
\end{quote}

For the ABC API baselines, the available reporting records do not establish immutable model snapshots or complete prompt and retry specifications. The model names in the results should therefore be read as reported service names. A complete accounting of billed cost and end-to-end latency is unavailable across the API runs. We report recognition accuracy, without making a cost or speed comparison between these services.

Legato 2 supplies published scores rather than prediction files, so its coverage is restricted to the four reported OMR-NED conditions.

\subsection{Music and text metrics}
\label{sec:metrics}
OMR-NED compares each target-page prediction with its corresponding page reference. Each domain sums the page-level notation edits and GT-plus-prediction notation sizes before division. The four-condition OpenScore macro is the equal-weight mean of domain OMR-NED values.

CER, SER and LER use the same deterministic XML-to-ABC conversion for prediction and reference. Their units are characters, lexical tokens including punctuation, and text lines. For each domain we divide the summed edit distance by the summed reference-unit count and multiply by 100. Scores can exceed 100 because insertions are not clipped. Their units are defined by this conversion.

TEDn compares notation trees derived from XML using rational durations. All predicted and reference parts enter the comparison, and the full authored measure sequences remain available to the metric. The edit cost for each page is divided by the deletion cost of its complete reference tree and multiplied by 100; the domain score averages these page percentages over its fixed sample count. An empty model output therefore contributes 100\%, while insertions can produce larger values. The tree representation excludes the lyric and layout fields measured separately. Scoring-resource failures remain unresolved until computation completes. Lyric-text CER is computed separately from lyric text extracted from XML and is distinct from ABC-based CER.

\subsection{Fixed-sample failure accounting}
All pages remain in their evaluation domain. Admissibility is metric-specific: for example, a prediction that supports notation comparison but not ABC conversion retains its OMR-NED and receives an empty prediction only for the affected text metrics. For empty predictions, notation edit counts equal the reference notation count and the prediction count is zero. For CER, SER and LER, all reference characters, lexical tokens or lines are deletions; the empty prediction contains zero lines. We aggregate these counts with successful pages before division. Successful error rates above 100\% remain unchanged.

Reference and evaluator failures remain separate from model errors. Table~\ref{tab:failures} gives the metric-specific counts.

\begin{table*}[!t]
\centering\footnotesize
\begin{tabular}{@{}lrrrrr@{}}
\toprule
Condition & $N$ & \multicolumn{2}{c}{Cecilia-OMR} & \multicolumn{2}{c}{GPT-6 Astra (XML)}\\
 & & Invalid & ABC-only & Invalid & ABC-only\\
\midrule
Rendered OpenScore String Quartets & 252 & 11 & 0 & 52 & 0\\
Camera OpenScore String Quartets & 252 & 22 & 0 & 56 & 0\\
Rendered OpenScore Lieder & 64 & 5 & 1 & 5 & 2\\
Camera OpenScore Lieder & 64 & 6 & 0 & 5 & 1\\
Rendered OpenScore Lieder with lyrics & 64 & 7 & 0 & 6 & 0\\
Camera OpenScore Lieder with lyrics & 64 & 13 & 0 & 5 & 0\\
CVC-MUSCIMA & 100 & 0 & 1 & 15 & 0\\
\bottomrule
\end{tabular}
\caption{Failure counts for notation and ABC-based text evaluation. Invalid denotes an unusable XML prediction under the notation evaluator. ABC-only denotes an additional model-structure failure in XML-to-ABC conversion whose notation score remains available. Counts record which predictions each metric can evaluate. All failure pages remain in the fixed evaluation sets.}
\label{tab:failures}
\end{table*}

The sequence metrics further distinguish content recovery from tree-level similarity. Their shared XML-to-ABC conversion exposes differences in characters, lexical units and lines under one deterministic representation. A low notation distance can therefore coexist with substantial line error when the predicted music is organised differently in the converted text. We report the conversion-based line error separately from the system-break F1 of final layout: the former concerns a symbolic text sequence, while the latter concerns where the rendered systems divide the score. Together, the two measures distinguish symbolic line structure from rendered system geometry.

\subsection{Final-layout metrics}
The reference and predicted MusicXML use the same MuseScore 3.2.3 renderer and fonts. Geometry is extracted from final SVG, positional exports and PDF page dimensions. Coordinates are normalised within the rendered page. The comparison measures agreement between the final layouts rendered from the two XML files.

System-break F1 compares system boundaries after alignment of logical measures. Page-break F1 compares page boundaries. Measures align by musical order; staffs align within matched systems; text words use normalised text and order. For an object category with $n_G$ reference and $n_P$ predicted objects, matched IoUs are summed and divided by $\max(n_G,n_P)$. Unmatched objects contribute zero. NPE uses centre displacement divided by the page diagonal, clipped to one; unmatched or cross-page objects contribute one, with the same object-count denominator. These rules include both matched geometry and omitted or extra objects.

Lyric-line initialisation is identical for all rendered predictions and references, including zero-length lyric segments. The notation, lyric content and geometric attributes are preserved before the common layout comparison.

The main geometric comparison evaluates Cecilia-OMR against Astra XML. It measures predicted layout instructions and automatic layout under one renderer. The other baselines enter the musical evaluation and qualitative comparison.

\section{Complete recognition and layout results}
Tables~\ref{tab:complete_secondary} and~\ref{tab:complete_geometry} give the numerical values and availability of results plotted in the main paper (Figure~\ref{Main-fig:metrics}). They preserve the complete domain sizes, common conversions and final-layout definitions described above. The geometric comparison is restricted to Cecilia-OMR and the direct-XML Astra condition, while all six systems with available predictions enter the secondary recognition table.

\section{Decoding strategy comparison}
The two strategies in Table~\ref{Main-tab:decoding_strategy} use the same Cecilia-OMR weights, ordered image windows, reference XML and output processing. Model parameters and the output projection use FP32, forward computation uses BF16 autocast, and TF32 is disabled. The comparison retains all 632 OpenScore benchmark pages, 128 Lieder with lyrics pages and 100 CVC-MUSCIMA pages. XML-availability selection uses the compiled candidate outputs to choose between unconstrained and grammar-prefix generation; reference XML enters only the subsequent evaluation. Both selected outputs receive the same deterministic recovery before evaluation.

\section{Predicted layout and automatic layout}
We compare two renderings of each Cecilia-OMR prediction on 48 fixed pages, with eight pages per condition. The first retains predicted layout instructions. The second removes page, system, staff and measure layout elements, explicit break and spacing attributes, measure widths, and position and curve-control coordinates. Both retain the same musical events, lyric attachments, part labels, scaling, fonts and appearance settings. An independent tree comparison verifies equality of all retained content. Both versions use the same MuseScore renderer and reference artifact. Table~\ref{Main-tab:layout_ablation} reports means over all 48 entries.

\section{Training-stage domain results}
\begin{table*}[!t]
\centering\small
\begin{tabular}{@{}lrrrrrr@{}}
\toprule
& \multicolumn{4}{c}{OpenScore benchmarks} & \multicolumn{2}{c}{Lieder with lyrics}\\
\cmidrule(lr){2-5}\cmidrule(l){6-7}
Stage & Rendered & Camera & Rendered & Camera & Rendered & Camera\\
 & Quartets & Quartets & Lieder & Lieder & & \\
\midrule
Base & .3615 & .5569 & .1308 & .4201 & .2478 & .5634\\
SFT & .2372 & .2359 & .2187 & .2336 & .2383 & .2732\\
GRPO & .2513 & .2307 & .1839 & .2270 & .2048 & .2862\\
\bottomrule
\end{tabular}
\caption{Domain-level OMR-NED for the training-stage comparison in Figure~\ref{Main-fig:scaling_demo}. Each Quartets condition contains 252 pages; each Lieder condition contains 64.}
\label{tab:model_variants}
\end{table*}

\section{Additional controlled evaluations}
\label{sec:additional_evaluations}
\subsection{Reward variants}
\label{sec:lyric_reward_ablation}
The reward variants initialise the policy and frozen reference from the same SFT model. Each uses 16 training groups, four candidates per group and seven optimiser updates. The original reward is $R_{\mathrm{original}}=0.95M+0.05F$, where $M$ is native notation accuracy and $F$ is format validity. The two additional variants use
\begin{align}
R_{\mathrm{lyric}}&=0.80M+0.15L+0.05F,\\
R_{\mathrm{format}}&=0.90M+0.10F.
\end{align}
The lyric variant applies to Lieder with lyrics; Quartets retain the original reward. Here, $L=\max(0,1-E_{\ell}/G_{\ell})$, with lyric character edits $E_{\ell}$ and reference character count $G_{\ell}$. Optimisation retains the original learning rate, temperature, training examples and group order. Each variant samples candidates from its current policy. Evaluation uses the same 48 target pages, with eight pages per condition. This small evaluation and short refinement support an exploratory comparison rather than a general conclusion about the best reward design.

Table~\ref{Main-tab:reward_main} reports the six notation scores. Table~\ref{tab:reward_lyrics} adds lyric-text CER for the two new variants. Each entry pools character edits and reference characters over eight pages.

\begin{table}[!htbp]
\centering\small
\begin{tabular}{@{}lrr@{}}
\toprule
Reward & Rendered (\%) & Camera (\%)\\
\midrule
Lyric & 12.2795 & 95.1220\\
Format & 11.7368 & 99.1516\\
\bottomrule
\end{tabular}
\caption{Lyric-text CER for the new GRPO reward variants. Each score pools eight pages; lower is better.}
\label{tab:reward_lyrics}
\end{table}

\subsection{Sampling temperature and candidate count}
\label{sec:temperature_candidates}
We compare temperatures $T\in\{0.3,0.7,1.0\}$ on 48 target pages, with eight pages from each of the six OpenScore conditions. Cecilia-OMR generates four independently seeded candidates at each temperature. A candidate pairs metadata and music from the same seed. The best-of-$N$ comparisons are nested: $N=1$ retains the first candidate, $N=2$ selects from the first two, and $N=4$ selects from all four.

Selection first prefers a complete pair with valid native XML, then maximises the mean temperature-adjusted log probability of its segment-count and token decisions. The denominator counts two unit-count decisions and all generated token decisions; BOS is excluded and EOS included. Ties retain the lower candidate index. If every candidate is invalid, the same probability rule selects an output whose invalid status is retained. Selection is performed separately at each temperature. The selected output receives the usual deterministic recovery. Reference notation is used only for evaluation. The generation budget permits up to 16 segments per unit and 8,192 tokens per segment; a sampled count above the segment cap produces an invalid candidate.

\begin{table*}[!t]
\centering\small
\setlength{\tabcolsep}{4pt}
\begin{tabular}{@{}lrrrrrrrrr@{}}
\toprule
& & \multicolumn{6}{c}{OMR-NED $\downarrow$} & \multicolumn{2}{c}{Lyric CER (\%) $\downarrow$}\\
\cmidrule(lr){3-8}\cmidrule(l){9-10}
$T$ & $N$ & RQ & CQ & RL & CL & RL+L & CL+L & Rendered & Camera\\
\midrule
$0.3$ & 1 & 0.3273 & 0.3881 & 0.1448 & 0.4248 & 0.2812 & 0.4434 & 46.68 & 72.96\\
$0.3$ & 2 & 0.3627 & 0.2999 & 0.1244 & 0.4662 & 0.2353 & 0.2400 & 39.82 & 68.61\\
$0.3$ & 4 & 0.3486 & 0.2342 & 0.1184 & 0.4277 & 0.2920 & 0.2863 & 44.64 & 71.05\\
$0.7$ & 1 & 0.3409 & 0.3580 & 0.2522 & 0.4111 & 0.3910 & 0.4259 & 48.91 & 65.64\\
$0.7$ & 2 & 0.3166 & 0.2763 & 0.1299 & 0.5651 & 0.4765 & 0.2777 & 52.65 & 75.50\\
$0.7$ & 4 & 0.3571 & 0.1438 & 0.0919 & 0.4854 & 0.3272 & 0.1968 & 19.00 & 74.23\\
$1.0$ & 1 & 0.4542 & 0.7486 & 0.4317 & 0.5280 & 0.6343 & 0.8120 & 59.91 & 96.71\\
$1.0$ & 2 & 0.3640 & 0.6352 & 0.1753 & 0.5308 & 0.5841 & 0.8250 & 57.19 & 99.15\\
$1.0$ & 4 & 0.3044 & 0.5394 & 0.1880 & 0.6805 & 0.4971 & 0.6342 & 56.78 & 90.99\\
\bottomrule
\end{tabular}
\caption{Sampling temperature and nested candidate-count results. Every domain contains the same eight target pages at all settings. RQ/CQ denote rendered/camera Quartets, RL/CL denote rendered/camera Lieder, and +L retains the vocal staff and lyrics. OMR-NED pools notation counts within each domain; lyric CER pools character counts. Candidate selection uses native XML validity and model probabilities, with references used only for evaluation.}
\label{tab:sampling_complete}
\end{table*}

The per-domain values in Table~\ref{tab:sampling_complete} supplement the macro comparison in Table~\ref{Main-tab:sampling_main}. All settings retain the same page and reference identities, with independent seeds reused across the nested candidate sets.

\section{The Figure 1 passage in three formats}
\label{sec:format_example}
The following listings give the complete passage displayed in Figure~\ref{Main-fig:representation}(a): the vocal part, measures~5--16, of Roger Quilter's \emph{The Night Piece}, together with its title and credit information. ABC, Cecilia and MusicXML describe the same twelve measures; these include the shorter examples in Figure~\ref{Main-fig:representation}(b). Cecilia and MusicXML retain the layout fields, while ABC expresses the musical events and lyrics. The final words ``of'' continue beyond the displayed passage in the original score. Long lines wrap for display.
\lstset{basicstyle=\ttfamily\fontsize{8}{10}\selectfont,breaklines=true,breakatwhitespace=false,literate={’}{{'}}1}

\subsection{ABC}
\begin{lstlisting}
X:1
T:To Julia, Op.8
T:The Night Piece
C:Roger Quilter
Z:Robert Herrick
Z:OpenScore (CC0)
L:1/8
M:2/4
I:linebreak $
K:C
V:1 treble nm="Clarinet in C"
V:1
 z2 z!p! E | A3/2 B/ A G | E2 E E | A3/2 B/ A G |$ %4
w: Her|eyes the glow- worm|lend thee, The|shoot- ing stars at-|
 E2 E E/ E/ | D D G G | D/ E/ D E G/ G/ |!<(! A c B G!<)! |$ %8
w: tend thee; And the|elves al- so, Whose|lit- tle eyes glow Like the|sparks of fire, the|
 (e>d c) G | (A>B A>B | G>A F>G | (E2 D)) A | %12
w: sparks, _ _ the|sparks _ _ _|_ _ _ _|* * of|

\end{lstlisting}

\subsection{Cecilia}
\begin{lstlisting}[firstline=2]
(|0K@e:4.0(|11(|12=#"To Julia, Op.8"))(|29=5)(|1O=#"The Night Piece")(|0L(|17=#"Transcribed from the IMSLP file(s) at https://imslp.org/wiki/Special:ReverseLookup/321919"@D:#"arranger")(|17=#"Roger Quilter"@D:#Z)(|17=#"Robert Herrick"@D:#"lyricist")(|1m=#"OpenScore (CC0)"))(|0M(|0N(|0O=6.7056)(|0P=40))(|0Q(|0R=1771.21)(|0S=1253.03)(|0D@D:#S(r=59.6517)(s=59.6517)(|0E=59.6517)(|0F=90.9091))(|0D@D:#T(r=59.6517)(s=59.6517)(|0E=59.6517)(|0F=90.9091)))(#"appearance"(#"line-widt
h"=1.8@D:#"light barline")(#"line-width"=5.5@D:#"heavy barline")(#"line-width"=5@D:#"beam")(#"line-width"=4.4@D:#1O)(#"line-width"=1.5@D:#1P)(#"line-width"=1@D:#"enclosure")(#"line-width"=1.1@D:#"ending")(#"line-width"=1@D:#"extend")(#"line-width"=1.6@D:#"leger")(#"line-width"=1.1@D:#2p)(#"line-width"=1.2@D:#"octave shift")(#"line-width"=2.1@D:#"slur middle")(#"line-width"=0.7@D:#"slur tip")(#"line-width"=1.1@D:#"staff")(#"line-width"=1.1@D:#"stem")(#"line-width"=2.1@D:#"tie middle")(#"line-widt
h"=0.7@D:#"tie tip")(#"line-width"=1@D:#"tuplet bracket")(#"line-width"=1.2@D:#"wedge")(#"note-size"=70@D:#U)(#"note-size"=70@D:#"grace")(#"note-size"=49@D:#"grace-cue"))(#"music-font"@N:#"Leland")(|0T@N:#"Edwin",O:10)(|0U@N:#"Edwin",O:10))(|0v@X:1(#"credit-type"=#"title")(|0t=#"To Julia, Op.8"@C:626.515152,B:1711.560455,O:14,Y:#P,Z:#R))(|0v@X:1(#"credit-type"=#"subtitle")(|0t=#"5. The Night Piece"@C:626.515152,B:1651.90882,O:20,Y:#P,Z:#R))(|0v@X:1(#"credit-type"=#"lyricist")(|0t=#"Robert Herric
k"@C:59.651667,B:1554.994599,Y:#r,Z:#V))(|0v@X:1(#"credit-type"=#Z)(|0t=#"Roger Quilter"@C:1193.378636,B:1554.994599,Y:#p,Z:#V))(|0v@X:1(#"credit-type"=#"rights")(|0t=#"OpenScore (CC0)"@C:626.515152,B:90.909091,O:9,Y:#P,Z:#V))(|0V(|0i@G:#"P1"(|0j=#"Clarinet in C"@R:#m)(6@G:#"P1-I1"(7=#"Clarinet")(#"instrument-sound"=#"wind.reed.clarinet"))(|0h@G:#"P1-I1")(8@G:#"P1-I1"(9=72))))(|0k@G:#"P1"[@A:5,E:332.3(w(|0l=4)(|02(|03=0))(1(2=2)(3=4))(|05(|06=#8)(|07=2)))(Y@K:#i(t(u(r=0)(s=0))(|00=125.05))){@C:6
1.36,B:-20!$4;&1;'3;}{@C:124.22,B:-20!$2;&1;'0;}<@I:#n(j(p@C:3.09,H:15(|0s)))(o@F:54.44)>{@C:62.11,B:-40:E4;$2;&1;'0;.1;/@C:6.5,B:-48.69,A:1,H:-30\o;_"Her";/}][@A:6,E:270.59{@C:-237.69:A4;$3;&1;'0;(U@C:34.55,B:-25).1;/@C:6.5,B:-48.69,A:1,H:-30\o;_"eyes";/}{@C:93.17:B4;$1;&1;'7;.2;/@C:6.5,B:-48.69,A:1,H:-30\o;_"the";/}{@C:41.40:A4;$2;&1;'0;.1;/@C:6.5,B:-48.69,A:1,H:-30\6;_"glow";/}{@C:62.11:G4;$2;&1;'0;.1;/@C:6.5,B:-48.69,A:1,H:-30\5;_"worm";/}][@A:7,E:260.24{@C:-196.68:E4;$4;&1;'3;.1;/@C:6.5,B:-
48.69,A:1,H:-30\o;_"lend";/}{@C:124.22:E4;$2;&1;'0;.1;/@C:8.81,B:-48.69,A:1,H:-30\o;_"thee,";/}{@C:62.11:E4;$2;&1;'0;.1;/@C:6.5,B:-48.69,A:1,H:-30\o;_"The";/}][@A:8,E:270.59{@C:-186.33:A4;$3;&1;'0;(U@C:34.55,B:-25).1;/@C:6.5,B:-48.69,A:1,H:-30\6;_"shoot";/}{@C:93.17:B4;$1;&1;'7;.1;/@C:6.5,B:-48.69,A:1,H:-30\5;_"ing";/}{@C:41.40:A4;$2;&1;'0;.1;/@C:6.5,B:-48.69,A:1,H:-30\o;_"stars";/}{@C:62.11:G4;$2;&1;'0;.1;/@C:6.5,B:-48.69,A:1,H:-30\6;_"at";/}][@A:9,E:323.99(Y@K:#i(t(u(r=0)(s=0))(|00=125.05))){@
C:61.36,B:-40:E4;$4;&1;'3;.1;/@C:6.5,B:-47.63,A:1,H:-30\5;_"tend";/}{@C:120.38:E4;$2;&1;'0;.1;/@C:9.02,B:-47.63,A:1,H:-30\o;_"thee;";/}{@C:60.19:E4;$1;&1;'7;.1;/@C:6.5,B:-47.63,A:1,H:-30\o;_"And";/}{@C:40.13:E4;$1;&1;'7;.1;/@C:6.5,B:-47.63,A:1,H:-30\o;_"the";/}][@A:10,E:264.46{@C:-260.17:D4;$2;&1;'0;.1;/@C:6.5,B:-47.63,A:1,H:-30\o;_"elves";/}{@C:60.20:D4;$2;&1;'0;.1;/@C:6.5,B:-47.63,A:1,H:-30\6;_"al";/}{@C:60.19:G4;$2;&1;'0;.1;/@C:9.02,B:-47.63,A:1,H:-30\5;_"so,";/}{@C:60.19:G4;$2;&1;'0;.1;/@C:6
.5,B:-47.63,A:1,H:-30\o;_"Whose";/}][@A:11,E:292.7{@C:-192.47:D4;$1;&1;'7;.1;/@C:6.5,B:-47.63,A:1,H:-30\6;_"lit";/}{@C:40.13:E4;$1;&1;'7;.1;/@C:6.5,B:-47.63,A:1,H:-30\5;_"tle";/}{@C:40.13:D4;$2;&1;'0;.1;/@C:6.5,B:-47.63,A:1,H:-30\o;_"eyes";/}{@C:60.19:E4;$2;&1;'0;.1;/@C:6.5,B:-47.63,A:1,H:-30\o;_"glow";/}{@C:60.19:G4;$1;&1;'7;.1;/@C:6.5,B:-47.63,A:1,H:-30\o;_"Like";/}{@C:40.13:G4;$1;&1;'7;.1;/@C:6.5,B:-47.63,A:1,H:-30\o;_"the";/}][@A:12,E:252.57<@I:#n(j(0@B:25,A:1,#"spread":13,D:#z))>{@C:-240.77
:A4;$2;&1;'0;.1;/@C:6.5,B:-47.63,A:1,H:-30\o;_"sparks";/}{@C:60.19:C5;$2;&1;'0;.2;/@C:6.5,B:-47.63,A:1,H:-30\o;_"of";/}{@C:60.20:B4;$2;&1;'0;.2;/@C:8.81,B:-47.63,A:1,H:-30\o;_"fire,";/}{@C:60.19:G4;$2;&1;'0;.1;/@C:6.5,B:-47.63,A:1,H:-30\o;_"the";/}<@I:#n(j(0@A:1,D:#g))>][@A:13,E:313.11(Y@K:#i(t(u(r=0)(s=0))(|00=125.05))){@C:61.36,B:-5:E5;$3;&1;'0;(U@C:77.85,B:-5).2;(I=#6@A:1)(N(l@#"bezier-x":28.28489,#"bezier-y":16.528239,A:1,D:#h))/@C:0.84,B:-48.4,A:1,H:-30\o;_"sparks,";(|0A)/}{@C:89.98:D5;$1;&
1;'7;.2;(I=#5@A:1)(I=#v@A:2)}{@C:39.99:C5;$2;&1;'0;.2;(N(l@#"bezier-x":-25.42371,#"bezier-y":20.660413,A:1,D:#g))}{@C:59.99:G4;$2;&1;'0;.1;/@C:6.5,B:-48.4,A:1,H:-30\o;_"the";/}][@A:14,E:273.79{@C:-239.28:A4;$3;&1;'0;(U@C:28.53,B:-25).1;(I=#6@A:1)(N(l@#"bezier-x":107.442625,#"bezier-y":38.184258,A:1,#"orientation":#"over",I:#n,D:#h))/@B:-48.4,A:1,H:-30\o;_"sparks";(|0A)/}{@C:89.98:B4;$1;&1;'7;.1;(I=#5@A:1)(I=#v@A:2)}{@C:39.99:A4;$3;&1;'0;(U@C:158.51,B:-25).1;(I=#6@A:1)}{@C:89.99:B4;$1;&1;'7;.1;(I
=#5@A:1)(I=#v@A:2)}][@A:15,E:275.07{@C:-218.68:G4;$3;&1;'0;(U@C:29.81,B:-25).1;(I=#6@A:1)}{@C:89.98:A4;$1;&1;'7;.1;(I=#5@A:1)(I=#v@A:2)}{@C:40.00:F4;$3;&1;'0;(U@C:159.79,B:-35).1;(I=#6@A:1)}{@C:89.98:G4;$1;&1;'7;.1;(I=#5@A:1)(I=#v@A:2)}][@A:16,E:271.75{@C:-223.28:E4;$4;&1;'3;.1;(N(l@#"bezier-x":28.229942,#"bezier-y":-20.35162,A:2,D:#h))}{@C:139.97:D4;$2;&1;'0;.1;(N(l@#"bezier-x":-103.85417,#"bezier-y":47.077243,A:1,D:#g)(l@#"bezier-x":-29.610089,#"bezier-y":-18.285533,A:2,D:#g))}{@C:59.99:A4;$2;
&1;'0;.1;/@C:6.5,B:-48.4,A:1,H:-30\o;_"of";/}]))

\end{lstlisting}

\subsection{MusicXML}
\begin{lstlisting}
<?xml version='1.0' encoding='utf-8'?>
<score-partwise version="4.0"><work><work-title>To Julia, Op.8</work-title></work><movement-number>5</movement-number><movement-title>The Night Piece</movement-title><identification><creator type="arranger">Transcribed from the IMSLP file(s) at https://imslp.org/wiki/Special:ReverseLookup/321919</creator><creator type="composer">Roger Quilter</creator><creator type="lyricist">Robert Herrick</creator><rights>OpenScore (CC0)</rights></identification><defaults
><scaling><millimeters>6.7056</millimeters><tenths>40</tenths></scaling><page-layout><page-height>1771.21</page-height><page-width>1253.03</page-width><page-margins type="even"><left-margin>59.6517</left-margin><right-margin>59.6517</right-margin><top-margin>59.6517</top-margin><bottom-margin>90.9091</bottom-margin></page-margins><page-margins type="odd"><left-margin>59.6517</left-margin><right-margin>59.6517</right-margin><top-margin>59.6517</top-margin><bottom-margin>90.9091</bottom-margin></p
age-margins></page-layout><appearance><line-width type="light barline">1.8</line-width><line-width type="heavy barline">5.5</line-width><line-width type="beam">5</line-width><line-width type="bracket">4.4</line-width><line-width type="dashes">1.5</line-width><line-width type="enclosure">1</line-width><line-width type="ending">1.1</line-width><line-width type="extend">1</line-width><line-width type="leger">1.6</line-width><line-width type="pedal">1.1</line-width><line-width type="octave shift">1.
2</line-width><line-width type="slur middle">2.1</line-width><line-width type="slur tip">0.7</line-width><line-width type="staff">1.1</line-width><line-width type="stem">1.1</line-width><line-width type="tie middle">2.1</line-width><line-width type="tie tip">0.7</line-width><line-width type="tuplet bracket">1</line-width><line-width type="wedge">1.2</line-width><note-size type="cue">70</note-size><note-size type="grace">70</note-size><note-size type="grace-cue">49</note-size></appearance><music-
font font-family="Leland" /><word-font font-family="Edwin" font-size="10" /><lyric-font font-family="Edwin" font-size="10" /></defaults><credit page="1"><credit-type>title</credit-type><credit-words default-x="626.515152" default-y="1711.560455" justify="center" valign="top" font-size="14">To Julia, Op.8</credit-words></credit><credit page="1"><credit-type>subtitle</credit-type><credit-words default-x="626.515152" default-y="1651.90882" justify="center" valign="top" font-size="20">5. The Night P
iece</credit-words></credit><credit page="1"><credit-type>lyricist</credit-type><credit-words default-x="59.651667" default-y="1554.994599" justify="left" valign="bottom">Robert Herrick</credit-words></credit><credit page="1"><credit-type>composer</credit-type><credit-words default-x="1193.378636" default-y="1554.994599" justify="right" valign="bottom">Roger Quilter</credit-words></credit><credit page="1"><credit-type>rights</credit-type><credit-words default-x="626.515152" default-y="90.909091"
 justify="center" valign="bottom" font-size="9">OpenScore (CC0)</credit-words></credit><part-list><score-part id="P1"><part-name print-object="no">Clarinet in C</part-name><score-instrument id="P1-I1"><instrument-name>Clarinet</instrument-name><instrument-sound>wind.reed.clarinet</instrument-sound></score-instrument><midi-device id="P1-I1" /><midi-instrument id="P1-I1"><midi-program>72</midi-program></midi-instrument></score-part></part-list><part id="P1"><measure number="5" width="332.3"><attri
butes><divisions>4</divisions><key><fifths>0</fifths></key><time><beats>2</beats><beat-type>4</beat-type></time><clef><sign>G</sign><line>2</line></clef></attributes><print new-system="yes"><system-layout><system-margins><left-margin>0</left-margin><right-margin>0</right-margin></system-margins><system-distance>125.05</system-distance></system-layout></print><note default-x="61.36" default-y="-20"><rest /><duration>4</duration><voice>1</voice><type>quarter</type></note><note default-x="185.58" d
efault-y="-20"><rest /><duration>2</duration><voice>1</voice><type>eighth</type></note><direction placement="above"><direction-type><dynamics default-x="3.09" relative-y="15"><p /></dynamics></direction-type><sound dynamics="54.44" /></direction><note default-x="247.69" default-y="-40"><pitch><step>E</step><octave>4</octave></pitch><duration>2</duration><voice>1</voice><type>eighth</type><stem>up</stem><lyric number="1" default-x="6.5" default-y="-48.69" relative-y="-30"><syllabic>single</syllab
ic><text>Her</text></lyric></note></measure><measure number="6" width="270.59"><note default-x="10.00" default-y="-25"><pitch><step>A</step><octave>4</octave></pitch><duration>3</duration><voice>1</voice><type>eighth</type><dot default-x="34.55" default-y="-25" /><stem>up</stem><lyric number="1" default-x="6.5" default-y="-48.69" relative-y="-30"><syllabic>single</syllabic><text>eyes</text></lyric></note><note default-x="103.17" default-y="-20"><pitch><step>B</step><octave>4</octave></pitch><dur
ation>1</duration><voice>1</voice><type>16th</type><stem>down</stem><lyric number="1" default-x="6.5" default-y="-48.69" relative-y="-30"><syllabic>single</syllabic><text>the</text></lyric></note><note default-x="144.57" default-y="-25"><pitch><step>A</step><octave>4</octave></pitch><duration>2</duration><voice>1</voice><type>eighth</type><stem>up</stem><lyric number="1" default-x="6.5" default-y="-48.69" relative-y="-30"><syllabic>begin</syllabic><text>glow</text></lyric></note><note default-x=
"206.68" default-y="-30"><pitch><step>G</step><octave>4</octave></pitch><duration>2</duration><voice>1</voice><type>eighth</type><stem>up</stem><lyric number="1" default-x="6.5" default-y="-48.69" relative-y="-30"><syllabic>end</syllabic><text>worm</text></lyric></note></measure><measure number="7" width="260.24"><note default-x="10.00" default-y="-40"><pitch><step>E</step><octave>4</octave></pitch><duration>4</duration><voice>1</voice><type>quarter</type><stem>up</stem><lyric number="1" default
-x="6.5" default-y="-48.69" relative-y="-30"><syllabic>single</syllabic><text>lend</text></lyric></note><note default-x="134.22" default-y="-40"><pitch><step>E</step><octave>4</octave></pitch><duration>2</duration><voice>1</voice><type>eighth</type><stem>up</stem><lyric number="1" default-x="8.81" default-y="-48.69" relative-y="-30"><syllabic>single</syllabic><text>thee,</text></lyric></note><note default-x="196.33" default-y="-40"><pitch><step>E</step><octave>4</octave></pitch><duration>2</dura
tion><voice>1</voice><type>eighth</type><stem>up</stem><lyric number="1" default-x="6.5" default-y="-48.69" relative-y="-30"><syllabic>single</syllabic><text>The</text></lyric></note></measure><measure number="8" width="270.59"><note default-x="10.00" default-y="-25"><pitch><step>A</step><octave>4</octave></pitch><duration>3</duration><voice>1</voice><type>eighth</type><dot default-x="34.55" default-y="-25" /><stem>up</stem><lyric number="1" default-x="6.5" default-y="-48.69" relative-y="-30"><s
yllabic>begin</syllabic><text>shoot</text></lyric></note><note default-x="103.17" default-y="-20"><pitch><step>B</step><octave>4</octave></pitch><duration>1</duration><voice>1</voice><type>16th</type><stem>up</stem><lyric number="1" default-x="6.5" default-y="-48.69" relative-y="-30"><syllabic>end</syllabic><text>ing</text></lyric></note><note default-x="144.57" default-y="-25"><pitch><step>A</step><octave>4</octave></pitch><duration>2</duration><voice>1</voice><type>eighth</type><stem>up</stem>
<lyric number="1" default-x="6.5" default-y="-48.69" relative-y="-30"><syllabic>single</syllabic><text>stars</text></lyric></note><note default-x="206.68" default-y="-30"><pitch><step>G</step><octave>4</octave></pitch><duration>2</duration><voice>1</voice><type>eighth</type><stem>up</stem><lyric number="1" default-x="6.5" default-y="-48.69" relative-y="-30"><syllabic>begin</syllabic><text>at</text></lyric></note></measure><measure number="9" width="323.99"><print new-system="yes"><system-layout>
<system-margins><left-margin>0</left-margin><right-margin>0</right-margin></system-margins><system-distance>125.05</system-distance></system-layout></print><note default-x="61.36" default-y="-40"><pitch><step>E</step><octave>4</octave></pitch><duration>4</duration><voice>1</voice><type>quarter</type><stem>up</stem><lyric number="1" default-x="6.5" default-y="-47.63" relative-y="-30"><syllabic>end</syllabic><text>tend</text></lyric></note><note default-x="181.74" default-y="-40"><pitch><step>E</s
tep><octave>4</octave></pitch><duration>2</duration><voice>1</voice><type>eighth</type><stem>up</stem><lyric number="1" default-x="9.02" default-y="-47.63" relative-y="-30"><syllabic>single</syllabic><text>thee;</text></lyric></note><note default-x="241.93" default-y="-40"><pitch><step>E</step><octave>4</octave></pitch><duration>1</duration><voice>1</voice><type>16th</type><stem>up</stem><lyric number="1" default-x="6.5" default-y="-47.63" relative-y="-30"><syllabic>single</syllabic><text>And</t
ext></lyric></note><note default-x="282.06" default-y="-40"><pitch><step>E</step><octave>4</octave></pitch><duration>1</duration><voice>1</voice><type>16th</type><stem>up</stem><lyric number="1" default-x="6.5" default-y="-47.63" relative-y="-30"><syllabic>single</syllabic><text>the</text></lyric></note></measure><measure number="10" width="264.46"><note default-x="21.89" default-y="-45"><pitch><step>D</step><octave>4</octave></pitch><duration>2</duration><voice>1</voice><type>eighth</type><stem
>up</stem><lyric number="1" default-x="6.5" default-y="-47.63" relative-y="-30"><syllabic>single</syllabic><text>elves</text></lyric></note><note default-x="82.09" default-y="-45"><pitch><step>D</step><octave>4</octave></pitch><duration>2</duration><voice>1</voice><type>eighth</type><stem>up</stem><lyric number="1" default-x="6.5" default-y="-47.63" relative-y="-30"><syllabic>begin</syllabic><text>al</text></lyric></note><note default-x="142.28" default-y="-30"><pitch><step>G</step><octave>4</oc
tave></pitch><duration>2</duration><voice>1</voice><type>eighth</type><stem>up</stem><lyric number="1" default-x="9.02" default-y="-47.63" relative-y="-30"><syllabic>end</syllabic><text>so,</text></lyric></note><note default-x="202.47" default-y="-30"><pitch><step>G</step><octave>4</octave></pitch><duration>2</duration><voice>1</voice><type>eighth</type><stem>up</stem><lyric number="1" default-x="6.5" default-y="-47.63" relative-y="-30"><syllabic>single</syllabic><text>Whose</text></lyric></note
></measure><measure number="11" width="292.7"><note default-x="10.00" default-y="-45"><pitch><step>D</step><octave>4</octave></pitch><duration>1</duration><voice>1</voice><type>16th</type><stem>up</stem><lyric number="1" default-x="6.5" default-y="-47.63" relative-y="-30"><syllabic>begin</syllabic><text>lit</text></lyric></note><note default-x="50.13" default-y="-40"><pitch><step>E</step><octave>4</octave></pitch><duration>1</duration><voice>1</voice><type>16th</type><stem>up</stem><lyric number
="1" default-x="6.5" default-y="-47.63" relative-y="-30"><syllabic>end</syllabic><text>tle</text></lyric></note><note default-x="90.26" default-y="-45"><pitch><step>D</step><octave>4</octave></pitch><duration>2</duration><voice>1</voice><type>eighth</type><stem>up</stem><lyric number="1" default-x="6.5" default-y="-47.63" relative-y="-30"><syllabic>single</syllabic><text>eyes</text></lyric></note><note default-x="150.45" default-y="-40"><pitch><step>E</step><octave>4</octave></pitch><duration>2<
/duration><voice>1</voice><type>eighth</type><stem>up</stem><lyric number="1" default-x="6.5" default-y="-47.63" relative-y="-30"><syllabic>single</syllabic><text>glow</text></lyric></note><note default-x="210.64" default-y="-30"><pitch><step>G</step><octave>4</octave></pitch><duration>1</duration><voice>1</voice><type>16th</type><stem>up</stem><lyric number="1" default-x="6.5" default-y="-47.63" relative-y="-30"><syllabic>single</syllabic><text>Like</text></lyric></note><note default-x="250.77"
 default-y="-30"><pitch><step>G</step><octave>4</octave></pitch><duration>1</duration><voice>1</voice><type>16th</type><stem>up</stem><lyric number="1" default-x="6.5" default-y="-47.63" relative-y="-30"><syllabic>single</syllabic><text>the</text></lyric></note></measure><measure number="12" width="252.57"><direction placement="above"><direction-type><wedge type="crescendo" default-y="25" spread="13" number="1" /></direction-type></direction><note default-x="10.00" default-y="-25"><pitch><step>A
</step><octave>4</octave></pitch><duration>2</duration><voice>1</voice><type>eighth</type><stem>up</stem><lyric number="1" default-x="6.5" default-y="-47.63" relative-y="-30"><syllabic>single</syllabic><text>sparks</text></lyric></note><note default-x="70.19" default-y="-15"><pitch><step>C</step><octave>5</octave></pitch><duration>2</duration><voice>1</voice><type>eighth</type><stem>down</stem><lyric number="1" default-x="6.5" default-y="-47.63" relative-y="-30"><syllabic>single</syllabic><text>
of</text></lyric></note><note default-x="130.39" default-y="-20"><pitch><step>B</step><octave>4</octave></pitch><duration>2</duration><voice>1</voice><type>eighth</type><stem>down</stem><lyric number="1" default-x="8.81" default-y="-47.63" relative-y="-30"><syllabic>single</syllabic><text>fire,</text></lyric></note><note default-x="190.58" default-y="-30"><pitch><step>G</step><octave>4</octave></pitch><duration>2</duration><voice>1</voice><type>eighth</type><stem>up</stem><lyric number="1" defau
lt-x="6.5" default-y="-47.63" relative-y="-30"><syllabic>single</syllabic><text>the</text></lyric></note><direction placement="above"><direction-type><wedge type="stop" number="1" /></direction-type></direction></measure><measure number="13" width="313.11"><print new-system="yes"><system-layout><system-margins><left-margin>0</left-margin><right-margin>0</right-margin></system-margins><system-distance>125.05</system-distance></system-layout></print><note default-x="61.36" default-y="-5"><pitch><s
tep>E</step><octave>5</octave></pitch><duration>3</duration><voice>1</voice><type>eighth</type><dot default-x="77.85" default-y="-5" /><stem>down</stem><beam number="1">begin</beam><notations><slur type="start" bezier-x="28.28489" bezier-y="16.528239" number="1" /></notations><lyric number="1" default-x="0.84" default-y="-48.4" relative-y="-30"><syllabic>single</syllabic><text>sparks,</text><extend /></lyric></note><note default-x="151.34" default-y="-10"><pitch><step>D</step><octave>5</octave><
/pitch><duration>1</duration><voice>1</voice><type>16th</type><stem>down</stem><beam number="1">end</beam><beam number="2">backward hook</beam></note><note default-x="191.33" default-y="-15"><pitch><step>C</step><octave>5</octave></pitch><duration>2</duration><voice>1</voice><type>eighth</type><stem>down</stem><notations><slur type="stop" number="1" bezier-x="-25.42371" bezier-y="20.660413" /></notations></note><note default-x="251.32" default-y="-30"><pitch><step>G</step><octave>4</octave></pit
ch><duration>2</duration><voice>1</voice><type>eighth</type><stem>up</stem><lyric number="1" default-x="6.5" default-y="-48.4" relative-y="-30"><syllabic>single</syllabic><text>the</text></lyric></note></measure><measure number="14" width="273.79"><note default-x="12.04" default-y="-25"><pitch><step>A</step><octave>4</octave></pitch><duration>3</duration><voice>1</voice><type>eighth</type><dot default-x="28.53" default-y="-25" /><stem>up</stem><beam number="1">begin</beam><notations><slur type="
start" orientation="over" placement="above" bezier-x="107.442625" bezier-y="38.184258" number="1" /></notations><lyric number="1" default-y="-48.4" relative-y="-30"><syllabic>single</syllabic><text>sparks</text><extend /></lyric></note><note default-x="102.02" default-y="-20"><pitch><step>B</step><octave>4</octave></pitch><duration>1</duration><voice>1</voice><type>16th</type><stem>up</stem><beam number="1">end</beam><beam number="2">backward hook</beam></note><note default-x="142.01" default-y=
"-25"><pitch><step>A</step><octave>4</octave></pitch><duration>3</duration><voice>1</voice><type>eighth</type><dot default-x="158.51" default-y="-25" /><stem>up</stem><beam number="1">begin</beam></note><note default-x="232.00" default-y="-20"><pitch><step>B</step><octave>4</octave></pitch><duration>1</duration><voice>1</voice><type>16th</type><stem>up</stem><beam number="1">end</beam><beam number="2">backward hook</beam></note></measure><measure number="15" width="275.07"><note default-x="13.32
" default-y="-30"><pitch><step>G</step><octave>4</octave></pitch><duration>3</duration><voice>1</voice><type>eighth</type><dot default-x="29.81" default-y="-25" /><stem>up</stem><beam number="1">begin</beam></note><note default-x="103.30" default-y="-25"><pitch><step>A</step><octave>4</octave></pitch><duration>1</duration><voice>1</voice><type>16th</type><stem>up</stem><beam number="1">end</beam><beam number="2">backward hook</beam></note><note default-x="143.30" default-y="-35"><pitch><step>F</
step><octave>4</octave></pitch><duration>3</duration><voice>1</voice><type>eighth</type><dot default-x="159.79" default-y="-35" /><stem>up</stem><beam number="1">begin</beam></note><note default-x="233.28" default-y="-30"><pitch><step>G</step><octave>4</octave></pitch><duration>1</duration><voice>1</voice><type>16th</type><stem>up</stem><beam number="1">end</beam><beam number="2">backward hook</beam></note></measure><measure number="16" width="271.75"><note default-x="10.00" default-y="-40"><pit
ch><step>E</step><octave>4</octave></pitch><duration>4</duration><voice>1</voice><type>quarter</type><stem>up</stem><notations><slur type="start" bezier-x="28.229942" bezier-y="-20.35162" number="2" /></notations></note><note default-x="149.97" default-y="-45"><pitch><step>D</step><octave>4</octave></pitch><duration>2</duration><voice>1</voice><type>eighth</type><stem>up</stem><notations><slur type="stop" number="1" bezier-x="-103.85417" bezier-y="47.077243" /><slur type="stop" number="2" bezier
-x="-29.610089" bezier-y="-18.285533" /></notations></note><note default-x="209.96" default-y="-25"><pitch><step>A</step><octave>4</octave></pitch><duration>2</duration><voice>1</voice><type>eighth</type><stem>up</stem><lyric number="1" default-x="6.5" default-y="-48.4" relative-y="-30"><syllabic>single</syllabic><text>of</text></lyric></note></measure></part></score-partwise>
\end{lstlisting}

    \clearpage
    \endgroup
  \fi
\fi
\end{document}